\documentclass[lettersize,journal]{IEEEtran}
\usepackage{amsmath,amsfonts}
\usepackage{algorithmic}
\usepackage{algorithm}

\usepackage{booktabs}
\usepackage{array}
\usepackage[caption=false,font=normalsize]{subfig}
\usepackage{textcomp}
\usepackage{stfloats}
\usepackage{url}
\usepackage{verbatim}
\usepackage{graphicx}
\usepackage{pifont}
\usepackage{cite}
\usepackage{multirow}
\begin{document}

\title{Multiobjective Vehicle Routing Optimization with Time Windows: A Hybrid Approach Using Deep Reinforcement Learning and NSGA-II}

\author{Rixin Wu, Ran Wang,~\IEEEmembership{Member,~IEEE}, Jie Hao,~\IEEEmembership{Member,~IEEE}, Qiang Wu,~\IEEEmembership{Member,~IEEE}, Ping Wang,~\IEEEmembership{Fellow,~IEEE}, and Dusit Niyato,~\IEEEmembership{Fellow,~IEEE}
	%\thanks{Manuscript received January 5, 2024. (Corresponding author: Ran Wang)}
	
	\thanks{Corresponding author: Ran Wang.}
	
	\thanks{Rixin Wu, Ran Wang, Jie Hao, and Qiang Wu are with the College of Computer Science and Technology, Nanjing University of Aeronautics and Astronautics, Nanjing 211106, China, and also with the Collaborative Innovation Center of Novel Software Technology and Industrialization, Nanjing, Jiangsu 210038, China (e-mail: wurixin@nuaa.edu.cn; wangran@nuaa.edu.cn; haojie@nuaa.edu.cn; wu.qiang@nuaa.edu.cn).}
	
	\thanks{Ping Wang is with the Department of Electrical Engineering and Computer Science, Lassonde School of Engineering, York University M3J 1P3, Canada (e-mail: pingw@eecs.yorku.ca).}
	
	\thanks{D. Niyato is with the School of Computer Science and Engineering, Nanyang Technological University 639798, Singapore (e-mail: dniyato@ntu.edu.sg).}
	
}

% The paper headers
% \markboth{Journal of \LaTeX\ Class Files,~Vol.~14, No.~8, August~2021}%
% {Shell \MakeLowercase{\textit{et al.}}: A Sample Article Using IEEEtran.cls for IEEE Journals}

% \IEEEpubid{0000--0000/00\$00.00~\copyright~2021 IEEE}
% Remember, if you use this you must call \IEEEpubidadjcol in the second
% column for its text to clear the IEEEpubid mark.

\maketitle

\begin{abstract}
This paper proposes a weight-aware deep reinforcement learning (WADRL) approach designed to address the multiobjective vehicle routing problem with time windows (MOVRPTW), aiming to use a single deep reinforcement learning (DRL) model to solve the entire multiobjective optimization problem. The Non-dominated sorting genetic algorithm-II (NSGA-II) method is then employed to optimize the outcomes produced by the WADRL, thereby mitigating the limitations of both approaches. Firstly, we design an MOVRPTW model to balance the minimization of travel cost and the maximization of customer satisfaction. Subsequently, we present a novel DRL framework that incorporates a transformer-based policy network. This network is composed of an encoder module, a weight embedding module where the weights of the objective functions are incorporated, and a decoder module. NSGA-II is then utilized to optimize the solutions generated by WADRL. Finally, extensive experimental results demonstrate that our method outperforms the existing and traditional methods. Due to the numerous constraints in VRPTW, generating initial solutions of the NSGA-II algorithm can be time-consuming. However, using solutions generated by the WADRL as initial solutions for NSGA-II significantly reduces the time required for generating initial solutions. Meanwhile, the NSGA-II algorithm can enhance the quality of solutions generated by WADRL, resulting in solutions with better scalability. Notably, the weight-aware strategy significantly reduces the training time of DRL while achieving better results, enabling a single DRL model to solve the entire multiobjective optimization problem.
\end{abstract}


\section{Introduction}

\IEEEPARstart{V}{ehicle} routing problem (VRP), a pivotal sector in transportation logistics, plays a crucial role in modern traffic management and operational efficiency. This significance is underscored by the fact that optimized vehicle routing can significantly reduce operational costs, a critical factor in the highly competitive and cost-sensitive transportation industry \cite{vrp1}. As the demand for precise delivery schedules escalates and the transportation sector becomes more competitive, incorporating customer delivery time windows has become not just a preference but a necessity. This evolution has catapulted the study of the vehicle routing problem with time windows (VRPTW) to the forefront of transportation research, marking it as a key area for innovative solutions in traffic and logistics management \cite{vrp2, vrp3}. The advancements in VRPTW are not just academic pursuits; they directly translate to more efficient, reliable, and cost-effective transportation systems, reinforcing the backbone of global trade and commerce.

In the study of VRPTW, time windows are typically categorized into two distinct types: hard time windows and soft time windows. In scenarios involving hard time windows, vehicles are obligated to adhere strictly to the scheduled delivery times. Specifically, for a VRP with a hard time window, the vehicle must arrive and conduct deliveries within the predefined time window. In instances where the vehicle arrives before the onset of the hard time window, it is requisite for the vehicle to wait until the specified start time before proceeding with the delivery. Conversely, if the vehicle arrives within the defined hard time window, it is authorized to proceed with the delivery directly. However, delivering goods subsequent to the expiration of the hard time window is strictly prohibited \cite{vrptw1, vrptw2}.	

In the VRP with soft time windows, there exists a permissible degree of flexibility for delivery vehicles to operate outside the designated time windows \cite{vrptw3, transformer4}. This flexibility, however, often incurs a consequential trade-off with customer satisfaction levels. An optimization strategy prioritizing the shortest path distance (or minimal cost) can result in the infringement of time windows for certain customers, thereby adversely affecting overall customer satisfaction \cite{vrptw4}. To enhance customer satisfaction, managers may find themselves compelled to augment the fleet size or extend the driving routes, which paradoxically escalates transportation costs. This scenario elucidates a prevalent conflict in VRPTW: the dichotomy between minimizing transportation costs and maximizing customer satisfaction. Consequently, the incorporation of multiobjective optimization emerges as a pivotal consideration within VRPTW.

In the domain of single-objective VRPTW (SOVRPTW) which predominantly focuses on minimizing travel costs, constraints such as time windows, vehicle capacity, and the number of available vehicles need to be considered simultaneously, hence significantly amplifying the complexity of solving SOVRPTW. Presently, the most commonly used algorithms for solving SOVRPTW are heuristic algorithms, such as genetic algorithm \cite{ga1}, particle swarm algorithm \cite{pso}, etc. The typical procedure for solving SOVRPTW using these algorithms involves the random generation of an initial set of solutions, followed by the application of heuristic strategies to enhance the quality of these solutions, thereby obtaining more cost-effective routes \cite{ga1}. Particularly in largerscale VRPTW instances, generating initial solutions that meet the constraints can be time-consuming, and the quality of these initial solutions is often low. These challenges become even more pronounced when dealing with multiobjective VRPTW (MOVRPTW), which simultaneously considers driving costs and customer satisfaction. Therefore, the development of an effective algorithm for solving MOVRPTW becomes an urgent and pressing need.

In this paper, we propose an MOVRPTW model which takes into account both travel cost and customer satisfaction in a holistic manner. To address this model, we develop a novel multiobjective optimization algorithm that combines weight-aware deep reinforcement learning (WADRL) and the non-dominated sorting genetic algorithm-II (NSGA-II) to solve MOVRPTW. The main contributions of this paper are summarized as follows:

\begin{itemize}[]
	\item  To mitigate the extensive training time typically associated with training individual subproblems in deep reinforcement learning (DRL) for multiobjective optimization, we have innovatively conceptualized and implemented a WADRL method. Specifically, during each training process, a set of random weights is generated to describe the relationship between the two objective functions, and a set of continuous weight combinations are adopted during the test process to obtain vehicle routes under different weight combinations. Such a method enables only one single DRL model to obtain the Pareto front of the entire multiobjective optimization problem.

	\item The implementation of the WADRL approach, while innovative and time-saving, may lead to suboptimal performance, as the solutions generated cannot assure Pareto optimal. To enhance solution quality and achieve an ideal Pareto front, we propose integrating the initial solutions generated by WADRL into the NSGA-II. This hybrid approach not only improves solution quality but also addresses the challenge of time-consuming initial solution generation often encountered by traditional heuristic algorithms when dealing with complex constraint problems.

	\item To the best of our knowledge, based on the existing literature, our study represents the first application of DRL in solving the MOVRPTW. The experimental results demonstrate that our method markedly surpasses existing traditional methods in terms of both convergence speed and solution diversity, particularly in the context of MOVRPTW. Concurrently, the implementation of a weight-aware strategy within the DRL framework significantly reduces training time and yields superior solutions.
\end{itemize}

The remainder of this paper is organized as follows. In Section II, related work is introduced. We present the system model and formulation of MOVRPTW in Section III. In Section IV, the details of WADRL combined with NSGA-II to solve multiobjective optimization problems are introduced. Simulation results and discussions are presented in Section V. Finally, this paper is concluded in Section VI.

\section{Related Work}

In recent years, there has been significant research focus on the vehicle routing problem with time windows (VRPTW), leading to the development of numerous algorithms aimed at solving this problem. These algorithms can be categorized into three main groups: exact algorithms \cite{pa1, pa2, pa3}, heuristic algorithms \cite{ga1, ga2, pso, ts, nsga, mopso, moead, mosa}, and learning-based algorithms \cite{modrl1, modrl2}.

Exact algorithms approach problems by rigorously deriving mathematical formulations, thus establishing mathematical models for the problem, and proposing algorithms to find optimal solutions based on mathematical principles. The authors in \cite{pa1} employed the column generation method to achieve the shortest paths in VRPTW. Furthermore, branch and bound \cite{pa2} as well as branch and price \cite{pa3} methods have also been applied in the context of VRPTW. However, it's worth noting that exact algorithms, which seek optimal solutions from all feasible solutions according to certain rules, face exponential growth in search space and computational complexity as the number of customers in VRPTW increases. Moreover, these algorithms tend to be less effective when dealing with multiobjective VRPTW (MOVRPTW).

Heuristic algorithms are the most commonly used methods in solving VRPTW. In solving single-objective VRPTW (SOVRPTW), common approaches encompass genetic algorithm (GA) \cite{ga1, ga2}, particle swarm algorithm (PSO) \cite{pso}, and tabu search (TS) algorithm \cite{ts}, among others. In \cite{ga1}, Ursani \textit{et al}. employed a local genetic algorithm to implement small-scale VRPTW targeting the shortest path and produced better solutions than most other heuristics methods. Another genetic algorithm, which was centered on the insertion heuristic, was proposed for the resolution of VRPTW, as documented in \cite{ga2}. The experimental results revealed the algorithm was able to find the solution in less time. Furthermore, both PSO with local search \cite{pso} and the TS algorithm \cite{ts} were also used to solve SOVRPTW.

In real-life scheduling optimization, increasing demands have rendered SOVRPTW less capable. Consequently, many researchers have delved into the study of MOVRPTW. Jaber Jemai \textit{et al}. employed the non-dominated sorting genetic algorithm-II (NSGA-II) to simultaneously optimize the minimization of both total travel distance and carbon dioxide emissions \cite{nsga}. In ambulance route optimization, NSGA-II and the multiobjective particle swarm optimization (MOPSO) algorithm were employed to jointly optimize the latest service completion times and the number of patients whose medical conditions deteriorated due to delayed medical services \cite{mopso}. Additionally, multiobjective evolutionary algorithm based on decomposition (MOEA/D) \cite{moead} and multiobjective simulated annealing (MOSA) algorithm \cite{mosa} have also been utilized in addressing MOVRPTW.

In the process of employing heuristic algorithms for solving MOVRPTW, it is common practice to begin by generating a set of initial solutions that adhere to the problem's constraints. Subsequently, a set of heuristic strategies, such as those found in genetic algorithms (GA), including selection, crossover, and mutation operations, in conjunction with non-dominated sorting strategies, is applied to obtain the Pareto front. This approach is advantageous due to its simplicity and its capability to yield high-quality solutions. However, it is worth noting that heuristic algorithms typically generate initial solutions using random strategies, and for MOVRPTW instances with numerous constraints, obtaining a feasible solution can be time-consuming. Moreover, the overall effectiveness of the algorithm is greatly influenced by the quality of the initial solutions \cite{ga3}. Additionally, whenever there is any modification to the problem's information, even minor changes, it necessitates rerunning the heuristic algorithm, which is known as the \textit{No Free Lunch Theorem} \cite{nofree}. Moreover, when the dimension of the problem is particularly large, these algorithms require a substantial number of iterations for overall updates and iterative searches to achieve relatively favorable results, leading to extended computation times \cite{modrl1}.

With the advancement of artificial intelligence (AI) technology, deep reinforcement learning (DRL) has also been employed to tackle multiobjective optimization problems. DRL was initially proposed for solving multiobjective travelling salesman problems (MOTSP) in \cite{modrl1}, in which the authors decomposed the MOTSP into several subproblems through weight combinations. They employed a sequence-to-sequence pointer network with two recurrent neural networks (RNNs) to train models for each subproblem. Additionally, a parameter transfer strategy was used in the initialization process of model parameters for adjacent subproblem training. In \cite{modrl2}, a similar approach was adopted, where the authors decomposed multiobjective vehicle routing problem (MOVRP) into multiple scalar subproblems based on weights. They utilized the pointer network to address each subproblem and trained the parameters of the policy network using the policy gradient algorithm of reinforcement learning to obtain the Pareto front of MOVRP.

The general method of deep reinforcement learning to solve multiobjective optimization problems is to convert the problem into multiple subproblems based on multiple sets of weight combinations with the same interval. Each subproblem can be regarded as a single objective optimization problem. Subsequently, a DRL-based model is trained for each subproblem, and only after training all models for the subproblems can the entire multiobjective optimization problem be addressed. While this method may be effective for biobjective optimization problems, as the number of objective functions increases, the training time escalates exponentially, which becomes impractical for solving it. Additionally, it is challenging to ensure that the DRL-based models for each subproblem are adequately trained, causing the solutions of some subproblems to be dominated by the solutions of other subproblems \cite{modrl1}, rendering these solutions meaningless in multiobjective optimization problems.

Considering the limitations of DRL in solving multiobjective optimization problems, we propose WADRL, which enables a single DRL model to address the entire multiobjective optimization problem. Furthermore, to overcome the drawbacks of heuristic algorithms in solving multiobjective optimization problems, we utilize the solutions generated by WADRL as the initial solutions of NSGA-II, ensuring that the initial solutions are feasible and of high quality. After undergoing evolution with NSGA-II, it is basically guaranteed that all solutions are Pareto optimal.

\section{Problem Statement}

\subsection{System Model}

The logistics businesses typically schedule service appointments with customers through phone calls or text messages before delivering goods or providing services. However, the delivery vehicles may arrive earlier or later than the scheduled time due to inefficient delivery routes or traffic jams. Customers usually have some level of tolerance for delays or early arrivals, but this may reduce customer satisfaction. Therefore, logistics companies should aim to improve customer satisfaction while simultaneously reducing travel cost \cite{moaco}.

\begin{figure}[!t]
	\centering
	\includegraphics[width=3in]{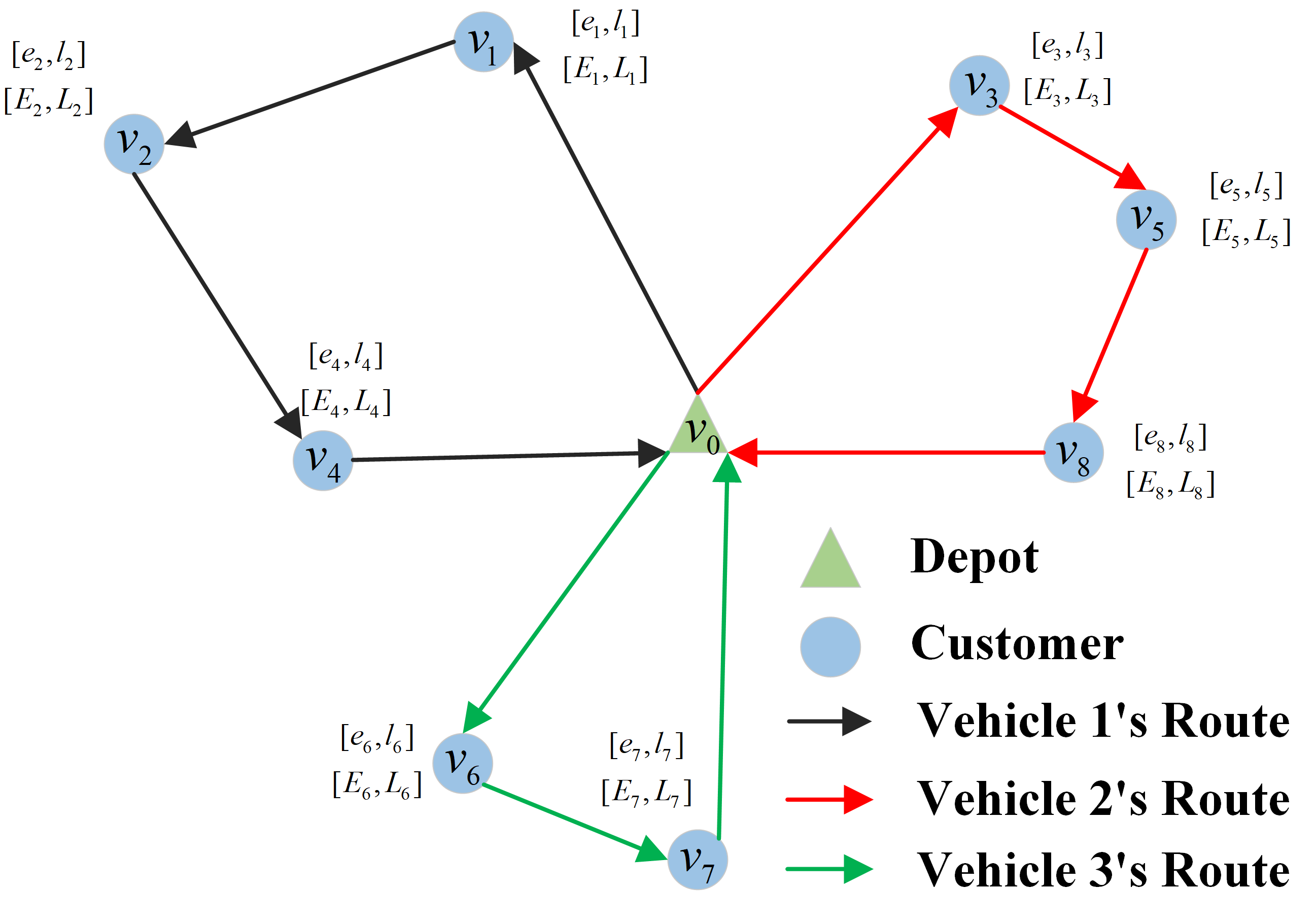}
	\caption{A MOVRPTW instance with three vehicles and eight customers.}
	\label{fig_1}
\end{figure}

We consider a multivehicle routing system in which $K$ vehicles are instructed to deliver goods to the set $H$ of customers. As illustrated in Fig. \ref{fig_1},  the scenario involves three vehicles tasked with delivering goods to eight customers. Our system is formally conceptualized as a graph $G=(V, E)$, where the set of vertices $V$ comprises a depot vertex $v_0$ and a set of customer vertices $C=\left\{v_i, i=1,2, \ldots, h\right\}$. The edges $E$ in the graph represent the roads connecting these vertices, each associated with a specific travel cost.

The coordinate of the depot is $u_0$, where each vehicle $k$ must depart from and return to. The information of each customer $v_i$ includes the coordinate $c_i$, the demand for goods $d_i$, the interval $\left[e_i, l_i\right]$ and the interval $\left[E_i, L_i\right]$. Among them, the interval  $\left[e_i, l_i\right]$ denotes the soft time window for customer $i$. Service provided to the customer within the intervals $\left[E_i, e_i\right]$ or $\left[l_i, L_i\right]$ leads to a decrease in customer satisfaction. The customer being served outside of the hard time interval $\left[E_i, L_i\right]$ is not permitted. The definition of the hard time window interval is typically as follows:

\begin{equation}
	E_i=e_i-\zeta_i^e\left(l_i-e_i\right),
\end{equation}

\begin{equation}
	L_i=l_i+\zeta_i^l\left(l_i-e_i\right),
\end{equation}
where $\zeta_i^e$ and $\zeta_i^l$ are the parameters used to define the allowable maximum violation. We define $S_i\left(a_i\right)$ as the satisfaction of customer $i$ when the vehicle arrives at customer $i$ at time $a_i$, and $S_i\left(a_i\right)$ can be defined as:

\begin{equation}
	S_i\left(a_i\right)= \begin{cases}0, & a_i<E_i \vee a_i>L_i \\ \frac{a_i-E_i}{e_i-E_i}, & E_i \leq a_i<e_i \\ 1, & e_i \leq a_i \leq l_i \\ \frac{L_i-a_i}{L_i-l_i}, & l_i<a_i \leq L_i\end{cases}.
\end{equation}

The key nomenclature used in this paper, along with their respective definitions, is outlined in TABLE \ref{tab:table1}.

\newcommand{\customrowheight}{1.2} % 自定义行高

\renewcommand{\arraystretch}{\customrowheight}
\begin{table}[!t]
	\caption{Nomenclature Used in This Paper\label{tab:table1}}
	\centering
	\begin{tabular}{lp{7cm}}
		\hline
		Symbol & Definition\\
		\hline

		$H^*$ &Set of customer vertices\\
	    $H$ & Set of customer and depot vertices, $H = H^* \cup \{0\}$ \\
		$K$ & Set of vehicles\\
	    $x_{i j k}$ & Whether vehicle $k$ travels from customer $i$ to customer $j$ \\
	    $y_{i j k}$ & Whether customer $i$ is served by vehicle $k$ \\
	    $c_k^1$ & Travel cost of vehicle $k$ per unit distance \\
	    $c_k^2$ &  Fixed cost generated by using vehicle $k$ \\
	    $a_i$ & Time when the vehicle arrives at customer $i$ \\
	   	\multirow{2}{*}{$S_i\left(a_i\right)$} & Satisfaction of customer $i$ when the vehicle arrives at \\
	    & customer $i$ at time $a_i$ \\
	    $\left[e_i, l_i\right]$ & Soft time window of customer $i$\\
	    $\left[E_i, L_i\right]$ & Hard time window of customer $i$ \\
	    $u_k$ & Capacity of vehicle $k$ \\
	    $w_i$ & Waiting time of customer $i$ \\
	    $v_i$ & Service time of customer $i$ \\
	    $t_{ij}$ & Time required to move from customer $i$ to customer $j$ \\
	    $dis_{ij}$ & Distance between customer $i$ and customer $j$ \\
		\hline
	\end{tabular}
\end{table}

\subsection{Problem Formulation}

In this section, we propose a multiobjective vehicle routing problem with time windows model that considers the cost incurred by travel and vehicle usage as well as customer satisfaction, as follows:

The decision variables in this paper are defined as follows:

\begin{equation}
	x_{i j k}=\left\{\begin{array}{l}
		1, \text { if vehicle } k \text { travels from customer } i \text { to } j \\
		0, \text { otherwise, }
	\end{array}\right.
\end{equation}

\begin{equation}
	y_{i k}=\left\{\begin{array}{l}
		1, \text { if customer } i \text { is served by vehicle } k \\
		0, \text { otherwise. }
	\end{array}\right.
\end{equation}

The objective functions in this paper are defined as follows:

Objective function 1: minimizing the total travel cost, which can be expressed as:

\begin{equation}\label{obj1}
	\min f_1=\sum_{k \in K} c_k^1 \sum_{i \in H} \sum_{j \in H} d i s_{i j} x_{i j k}+\sum_{k \in K} c_k^2 \sum_{j \in H^*} x_{0 j k},
\end{equation}
where $c_k^1$ represents the travel cost of vehicle $k$ per unit distance, $c_k^2$ denotes the fixed cost associated with the utilization of vehicle $k$, and $dis_{ij}$ represents the distance between customer $i$ and customer $j$.

Objective function 2: maximizing the average customer satisfaction, which can be expressed as:

\begin{equation}\label{obj2}
	\max f_2=\frac{1}{h} S_i\left(a_i\right),
\end{equation}
where $a_i$ represents the arrival time of the vehicle at the location of customer $i$. Furthermore, $S_i\left(a_i\right)$ is defined as the satisfaction of customer $i$ when the vehicle arrives at customer $i$ at time $a_i$.

Subject to:

\begin{equation} \label{con1}
	\sum_{i \in H^*} d_{i} y_{i k}<u_k \quad \forall k \in K
\end{equation}

\begin{equation} \label{con2}
	\sum_{k \in K} y_{i k}=1 \quad \forall i \in H^*
\end{equation}

\begin{equation} \label{con3}
	\sum_{j \in H^*} x_{0 j k}-\sum_{i \in H^*} x_{i 0 k}=0 \quad \forall k \in K
\end{equation}

\begin{equation} \label{con4}
	\sum_{i \in H} x_{i j k}=y_{j k} \quad \forall k \in K, \forall j \in H^*
\end{equation}

\begin{equation} \label{con5}
	\sum_{j \in H} x_{i j k}=y_{i k} \quad \forall k \in K, \forall i \in H^*
\end{equation}

\begin{equation} \label{con6}
	w_0=v_0=0
\end{equation}

\begin{equation} \label{con7}
	w_i=\max \left\{0, E_i-a_i\right\} \quad \forall i \in H^*
\end{equation}

\begin{equation} \label{con8}
	E_i \leq a_i+w_i \leq L_i \quad \forall i \in H^*
\end{equation}

\begin{equation} \label{con9}
	a_j=\sum_{k \in K} \sum_{i \in H} x_{i j k}\left(a_i+w_i+v_i+t_{i j}\right) \quad \forall j \in H^*
\end{equation}

\begin{equation} \label{con10}
	x_{i j k} \in\{0,1\}, y_{i k} \in\{0,1\}, a_i \geq 0 \quad \forall i \in H, j \in H, k \in K
\end{equation}

In the aforementioned mathematical model, let $H^*$ represent the set of customer vertices, $H$ represent the combined set of customer and depot vertices, defined as $H = H^* \cup \{0\}$. The set of vehicles is denoted by $K$. The constraint (\ref{con1}) ensures that the demand of customers does not exceed the capacity of each vehicle, where $u_k$ represents the capacity of vehicle $k$. The constraint (\ref{con2}) mandates that each customer is served by exactly one vehicle. The constraint (\ref{con3}) guarantees that each vehicle must start and end its route at the depot. The constraints (\ref{con4}) and (\ref{con5}) ensure that each customer is served exactly once. The constraint (\ref{con6}) defines the waiting and service times of the depot, with $w_i$ and $v_i$ denoting the waiting and service times of customer $i$, respectively. The constraint (\ref{con7}) defines the waiting time. The constraint (\ref{con8}) enforces the hard time window requirements. Finally, the constraint (\ref{con9}) defines the relationship between the arrival time of a customer and the arrival time of the previous customer, and $t_{ij}$ represents the time required to travel from customer $i$ to customer $j$.

\section{Solutions and Algorithms}

Considering the complexity of multiobjective vehicle routing problem with time windows (MOVRPTW), particularly in scenarios involving a large number of customers, existing scheduling methods seem insufficient to effectively solve the problem. Therefore, this section introduces an innovative approach that amalgamates a weight-aware deep reinforcement learning (WADRL) methodology with the non-dominated sorting genetic algorithm-II (NSGA-II). We design this hybrid method to tackle the challenges posed by MOVRPTW.  Initially, the WADRL algorithm is employed to generate the initial population for the NSGA-II algorithm. Subsequently, the NSGA-II algorithm undertakes the task of optimizing this initial population, thus enhancing the solution's efficacy.

\subsection{General Framework}
\begin{figure}[!t]
	\centering
	\includegraphics[width=3in]{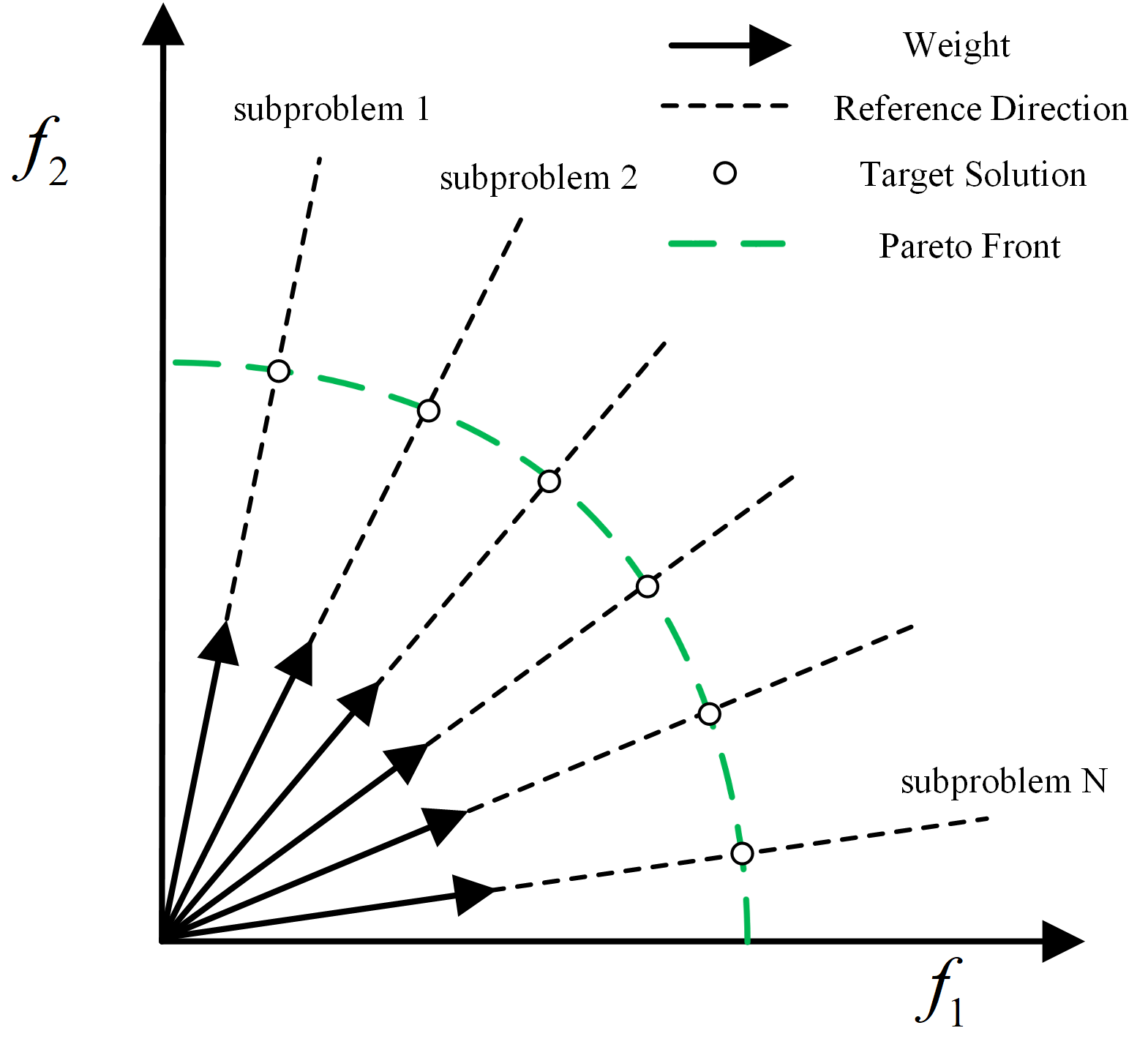}
	\caption{The decomposition strategy used in DRL for mops.}
	\label{fig_2}
\end{figure}

In traditional deep reinforcement learning (DRL) methods for solving multiobjective optimization problems, as illustrated in Fig. \ref{fig_2}, a biobjective optimization problem is decomposed into $N$ subproblems according to $N$ weight combinations. For the solution of each subproblem, the training of a DRL model is required. Consequently, a total of $N$ DRL models need to be trained to solve the biobjective optimization problem \cite{modrl1}. This approach appears to be effective for solving biobjective optimization problems. However, as the number of objective functions increases, the time required to solve the entire multiobjective optimization problem will increase exponentially.

Considering the shortcomings of the existing DRL methods to solve multiobjective optimization problems, our proposed algorithm takes a different approach. Similarly, we decompose the entire multiobjective optimization problem into $N$ subproblems based on $N$ weight combinations. During the training process of DRL, the algorithm randomly selects a weight combination for training each time. Furthermore, during the testing phase, the algorithm generates the corresponding optimal solutions for these $N$ subproblems. This approach enables a single DRL model to solve the entire multiobjective optimization problem, significantly reducing the model training time, particularly for many-objective optimization problems. The WADRL framework is drawn in Fig. \ref{fig_3}, in which a novel transformer architecture is also adopted.

In the original NSGA-II algorithm, the initial population is typically generated through a random process, and if the solution does not satisfy the constraints, it will be regenerated. In MOVRPTW, it is obviously impossible to generate solutions that satisfy constraints through a completely random method. While there exist methods to expedite the initial solution generation, they are still relatively inefficient \cite{vrptw5}, and the quality of these initial solutions is generally low. In our proposed algorithm, solutions generated by WADRL are used as the initial solutions for the NSGA-II algorithm. NSGA-II is then applied to optimize these solutions. The advantage of this approach is that the initial solutions generated by WADRL are guaranteed to satisfy the constraints. Furthermore, the generation of these initial solutions adheres to a carefully designed decomposition strategy. This approach guarantees a relatively uniform distribution across the Pareto front of the initial solutions. Importantly, this strategy not only facilitates coverage but also significantly enhances the quality of the initial solutions.

\subsection{The Markov Decision Process of MOVRPTW in Weight-aware Deep Reinforcement Learning}

Given the proficiency of reinforcement learning in managing sequential decision-making problems, we employ deep reinforcement learning to address the multiobjective vehicle routing problem with time windows (MOVRPTW). In this approach, MOVRPTW is modeled as a markov decision process (MDP), characterized by specific components: state $S$, action $A$, state transition function, and reward $R$. These elements are defined as follows.

\subsubsection{State}

	At any given step $t$, the state of the system encompasses three primary components: the vehicle state $v^t$, customer state $c^t$, and weight state $w^t$, defined as follows. The vehicle state $v^t$ is characterized by two dynamic attributes: its load, denoted as $u^t$, and the traveled time, symbolized as $\tau^t$. The term `dynamic' implies that these states may change as actions occur. The customer states encompass static states, including soft and hard time windows $e_i^t$, $l_i^t$, $E_i^t$ and $L_i^t$, coordinate $c_i^t$, and service time $v_i^t$. Static states indicate that these states will not change with the occurrence of actions. The dynamic state of a customer is demand $d_i^t$. The weight state is static, signifying the weights of two objective functions $w_1^t$ and $w_2^t$.
	
\subsubsection{Action}
	
	The action at each step	$t$, denoted as $a^t$, represents the next vertex to which the vehicle will travel. The entire sequence of actions from the initial step $0$ to the final step $T$ is represented as $\left\{a^0, a^1, a^2 \ldots, a^T\right\}$. $a^t \in\{0,1, \ldots, h\}$, where $h$ is the total number of vertices in the system. $a^t = 0$ means that the vehicle returns to the depot. $a^t \in\{1,2, \ldots, h\}$ signifies that the vehicle travels towards the customer vertex. The vehicle is initially at the depot vertex, i.e. $a^0 = 0$.
	
\subsubsection{State transition function}

	The current system state $S^t$ will transition to the next state $S^{t+1}$ based on the currently executed action $a^t$. All static states remain unchanged, while dynamic states may change. Once a customer node is visited, the demand of the customer becomes zero, which can be expressed as:
	
	\begin{equation}
		d_i^t=\left\{\begin{array}{l}
			0, \text { if } a^t=i \\
			d_i^{t-1}, \text { otherwise }.
		\end{array}\right.
	\end{equation}
	
	In addition, the state of the vehicle will also change. If the vehicle travels to a customer node, its load will decrease accordingly. Otherwise, if the vehicle travels to the depot node, its load will be replenished, as given by:
	\begin{equation}
		u^t=\left\{\begin{array}{l}
			u^{t-1}-d\left(a^t\right), \text { if } a^t \neq 0 \\
			u^0, \text { if } a^t=0.
		\end{array}\right.
	\end{equation}
	
	The traveled time is determined by the travel time between two vertices and previous traveled time, as given as follows:
	\begin{equation}
		\tau^t=\left\{\begin{array}{l}
			\max \left(\tau^{t-1}, E\left(a^{t-1}\right)\right) \\
			+v\left(a^{t-1}\right)+t\left(a^{t-1}, a^t\right), \text { if } a^t \neq 0 \\
			0, \text { if } a^t=0,
		\end{array}\right.
	\end{equation}
	where $E\left(a^{t-1}\right)$ denotes the earliest service start time of the customer $a^{t-1}$, $v\left(a^{t-1}\right)$ represents the service time of the customer $a^{t-1}$, and $t\left(a^{t-1}, a^t\right)$ is the travel time between the two vertices. The vehicle returning to the depot means dispatching a new vehicle and updating the traveled time.

\subsubsection{Reward}
	
	By executing actions $\left\{a_0, a_1, a_2 \ldots, a_T\right\}$, the traveled paths of all vehicles, as well as the arrival time and satisfaction of all customers, can be obtained. The objective function values can be calculated according to Eq. (\ref{obj1}) and Eq. (\ref{obj2}), and the total reward $R$ can be calculated using Eq. (\ref{reward}):
	\begin{equation} \label{reward}
		R=\left\{\begin{array}{l}
			-1000, \text { if } \sum_{i \in N^*} d_i^T \neq 0 \\
			-w_1 \frac{f_1-f_1^{\min }}{f_1^{\max }-f_1^{\min }}+w_2 \frac{f_2-f_2^{\min }}{f_2^{\max }-f_2^{\min }}, \text { otherwise },
		\end{array}\right.
	\end{equation}
	where $R = -1000$ indicates that after all vehicles have delivered the goods and returned to the depot, there are still customer vertices that have not been visited, which means the constraints are not met. Otherwise, the minus sign before the first objective function means that the objective function is a minimization function. $f_1^{min}$, $f_1^{max}$, $f_2^{min}$ and $f_2^{max}$ denote the minimum and maximum values of the two objective functions respectively, which aim to normalize the objective function. These values can be obtained using single objective DRL model, objective function structures or the existing research.

\subsection{The Policy network in Weight-aware Deep Reinforcement Learning}
\begin{figure*}[!t]
	\centering
	\includegraphics[width=6.5in]{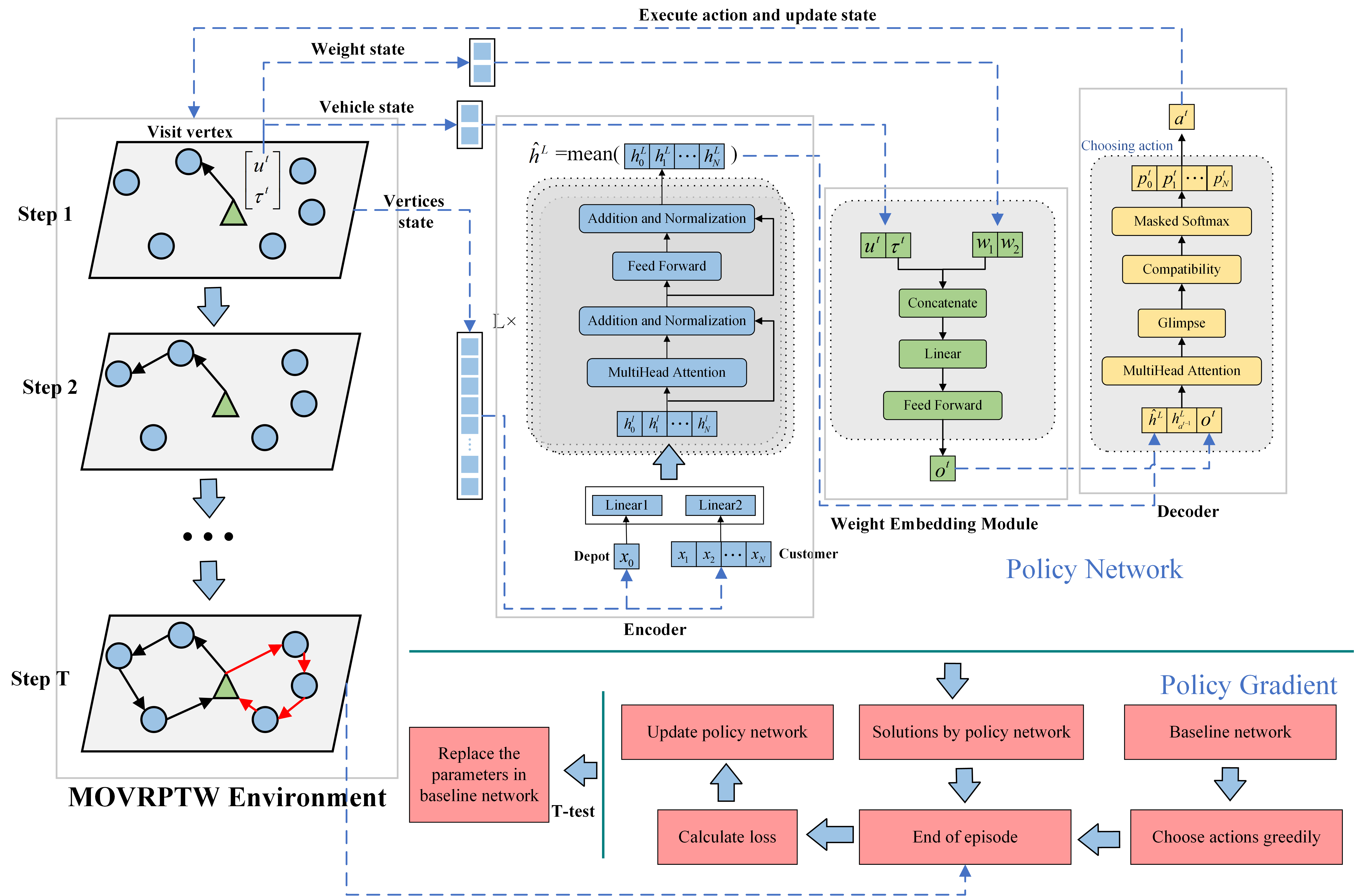}
	\caption{The framework of WADRL method.}
	\label{fig_3}
\end{figure*}

In our system model, various information needs to be considered simultaneously, such as customer coordinates, demand, time windows, and the weights of the objective functions. Therefore, learning-based methods need to be carefully designed to process the information. However, simple neural networks or learning strategies often struggle to handle the complex information mentioned above. Meanwhile, transformer architecture, as a type of self-attention mechanism, has been proven to perform well in many fields, including natural language processing (NLP) \cite {transformer1}, computer vision (CV) \cite {transformer2}, and recommender systems \cite {transformer3}. With the development of transformer technology, it has found wide applications in deep reinforcement learning, displaying excellence in problems like path planning \cite{transformer4}, the knapsack problem \cite{transformer5}, and reservoir operation \cite{transformer6}. Its advantage is that the attention mechanism in the transformer can effectively extract information through key-value-query maps.

In this paper, we employ the transformer architecture \cite{alg1} to model the delivering agent for solving MOVRPTW, as illustrated in Fig. \ref{fig_3}. The model primarily comprises an encoder module, a weight embedding module, and a decoder module \cite{alg1}. In the encoder process, the encoder embeds the original information of MOVRPTW into a high-dimensional space and utilizes a self-attention mechanism to extract features of the problem. In the weight embedding process, the embedding module encapsulates the state of vehicles and the weights associated with the two objectives of MOVRPTW. In the decoder process, the decoder generates vertex probabilities based on contextual information.

\subsubsection{The information encoder process}
	
	Initially, the encoder employs a linear layer to transform the features of the vertices (including both the depot and customers) into a high-dimensional space. This transformation results in the generation of initial embedding information, represented as $h^{(0)}=\left\{h_0^{(0)}, h_1^{(0)}, \ldots, h_N^{(0)}\right\}$, where $h_i^{(0)}$ corresponds to the initial embedded representation of vertex $i$. We employ separate linear projections to handle distinct information of warehouse and customer vertices, as follows \cite{alg2}:
	\begin{equation} \label{encoder1}
		h_i^{(0)}=\left\{\begin{array}{l}
			W_0^x\left[c_0, L_0\right]+b_0^x, \text { if } i=0 \\
			W_i^x\left[c_i, E_i, e_i, l_i, L_i, d_i, v_i\right]+b_i^x, \text { if } i \neq 0,
		\end{array}\right.
	\end{equation}
	where the depot vertex only needs to embed coordinate and latest arrival time information, while the customer vertices need to embed coordinate, soft and hard time window, demand, and service time information. The operation $[\cdot, \cdot, \ldots, \cdot]$ concatenates the information of the same dimension together. $W_0^x$, $b_0^x$, $W_i^x$ and $b_i^x$ are trainable linear projection parameters, and $W_0^x \in \mathbb{R}^{d_h \times 2}$, $b_0^x \in \mathbb{R}^{d_h}$, $W_i^x \in \mathbb{R}^{d_h \times 8}$, $b_i^x \in \mathbb{R}^{d_h}$, where $d_h$ represents the dimension of $h^{(0)}$. In our algorithm $d_h = 128$.
	
	Subsequent to the initial embedding, the encoded information undergoes processing through $L$ identical layers to yield the final embedding. Each of these layers is composed of several key components: multi-head attention layer (MHA), skip connection layer, batch normalization (BN) layer, and fully connected feedforward (FF) layer. The input of layer $l$ is the output of the preceding layer, i.e. $h^{(l-1)}=\left\{h_0^{(l-1)}, h_1^{(l-1)}, \ldots, h_N^{(l-1)}\right\}$. The MHA layer employs $M$ heads with a dimensionality of $d_k$. In our algorithm, we set $L=3$, $M = 8$ and $d_k=\frac{d_h}{M}=16$. For each head $m \in\{1,2, \ldots, M\}$, the query, key, and value are computed as follows \cite{alg3}:
	
	\begin{equation} \label{encoder2}
		Q_m^{(l)}=W_{Q m}^{(l)} h^{(l-1)},
	\end{equation}
	
	\begin{equation}  \label{encoder3}
		K_m^{(l)}=W_{K m}^{(l)} h^{(l-1)},
	\end{equation}
	
	\begin{equation} \label{encoder4}
		V_m^{(l)}=W_{V m}^{(l)} h^{(l-1)},
	\end{equation}
	where $W_{Q m}^{(l)}, W_{K m}^{(l)} \in \mathbb{R}^{d_k \times d_l}$, and $W_{V m}^{(l)} \in \mathbb{R}^{d_h \times d_h}$. Then the attention value $A_m^{(l)}$ and $M H A\left(h^{(l-1)}\right)$ are calculated as:
	\begin{equation} \label{encoder5}
		A_m^{(l)}=\operatorname{softmax}\left(\frac{Q_m^{(l)}\left(K_m^{(l)}\right)^T}{\sqrt{d_k}}\right) V_m^{(l)},
	\end{equation}
	
	\begin{equation} \label{encoder6}
		M H A\left(h^{(l-1)}\right)=\left[A_1^{(l)}, A_2^{(l)}, \ldots, A_M^{(l)}\right] W_O^{(l)}.
	\end{equation}
	
	$\hat{h}^{(l)}$ is obtained from the attention value through skip connection layer and BN layer, which can be expressed as follows:
	
	\begin{equation} \label{encoder7}
		\hat{h}^{(l)}=B N^{(l)}\left(h^{(l-1)}+M H A\left(h^{(l-1)}\right)\right).
	\end{equation}
	
	Ultimately, the output of the node embedding at layer $l$ is denoted as $h^{(l)}$, which can be obtained through an FF layer and a skip connection layer as follows:
	\begin{equation} \label{encoder8}
		h^{(l)}=B N^{(l)}\left(\hat{h}^{(l-1)}+F F\left(\hat{h}^{(l-1)}\right)\right).
	\end{equation}
	
\subsubsection{The weight embedding process}	
	
	As MOVRPTW is modeled as an MDP, each vehicle is treated as an agent. When the agent selects actions, it needs to consider the weights of the two objective functions in MOVRPTW. Hence, we incorporate a specialized weight embedding module within our framework. This module is carefully designed to capture the current state of the vehicle and the state of the weights associated with the objective functions. The strategy allows the agent to focus on the weights of the objective functions when making decisions. The output of the weight embedding module is defined as follows:
	
	\begin{equation} \label{weight1}
		o^t=F F\left(W_o\left[u^t, \tau^t, w_1, w_2\right]+b_o\right).
	\end{equation}

\subsubsection{The tour decoder process}	

	At each given step $t$, the agent selects a decision $a^t$ based on the current state $s^t$, which contains the embedding of the entire graph denoted as $\hat{h}^{(L)}$, the embedding information of the vertex corresponding to the vehicle's current location $h_{a^{t-1}}^{(L)}$ and the output $o^t$ of the weight embedding module as follows:
	
	\begin{equation} \label{decoder1}
		s^t=\left[\hat{h}^{(L)}, h_{a^{t-1}}^{(L)}, o^t\right].
	\end{equation}
	
	Subsequently, we calculate the contextual information of $s^t$ through the MHA layer, defining the query vector as the embedding of the agent, and the key vector and value vector as the embedding of customers: 
	\begin{equation} \label{decoder2}
		Q^{d t}=W_Q^d s^t,
	\end{equation}
	
	\begin{equation} \label{decoder3}
		K^d=W_K^d h^L,
	\end{equation}
	
	\begin{equation} \label{decoder4}
		V^d=W_V^d h^L.
	\end{equation}
	
	The compatibility between each vertex and the vehicle is calculated through an attention mechanism:
	
	\begin{equation} \label{decoder5}
		\lambda_t=C \cdot \tanh \left(\frac{\left(Q^{d t}\right)^T K^d}{\sqrt{d_k}}\right).
	\end{equation}
	
		\begin{algorithm}[!t]
		\renewcommand{\algorithmicrequire}{\textbf{Input:}}
		\renewcommand{\algorithmicensure}{\textbf{Output:}}
		\caption{The Training Process of Weight-Aware Deep Reinforcement Learning}
		\label{alg1} 
		\begin{algorithmic}[1]
			\REQUIRE
			Batch size and training epoch
			\ENSURE 
			The trained parameters $\theta$
			\STATE Initialization $\theta$ for policy network $\pi^{\theta}$, $\theta^{BL}$ for baseline network $\pi^{BL}$;
			\STATE Generate an MOVRPTW instance randomly;
			\FOR{each epoch}
			\FOR{each batch}
			\FOR{each step $t$}
			\STATE Calculate vertex embedding in Eqs.(\ref{encoder1})-(\ref{encoder8});
			\STATE Calculate weight embedding in Eq.(\ref{weight1});
			\STATE Calculate vertex probability vector in Eqs.(\ref{decoder1})-(\ref{decoder7});
			\STATE Choose action $a^t$;
			\IF{All available vehicles are returned to the depot vertex \textbf{or} all customer vertices are visited}
			\STATE Break;
			\ENDIF
			\ENDFOR
			\STATE Calculate the reward $R\left(\pi^\theta\right)$ and $R\left(\pi^{B L}\right)$ in Eq.(\ref{reward});
			\STATE Calculate $\nabla_\theta L(\theta)$;
			\STATE Update the parameters $\theta$;
			\ENDFOR
			\IF{T-Test result $<$ 0.95}
			\STATE $\theta^{B L} \leftarrow \theta$;
			\ENDIF
			\ENDFOR
		\end{algorithmic}
	\end{algorithm}
	
	In MOVRPTW, various constraints are established, meaning that at each step $t$, certain vertices are inaccessible due to these constraints. To navigate the diverse constraints inherent in the system, we also implement a masking rule. This rule functions to selectively exclude vertices that are not feasible for visitation during the current step. The primary masking rules include the following: \ding{172} Apart from the depot vertex, other previously visited vertices are masked; \ding{173} Customer vertices with demand exceeding the load of vehicle are masked; \ding{174} Vertices where the vehicle's arrival time at the node exceeds the hard time window are masked. Whether the vertex $i$ is masked at step $t$ is defined as:
	\begin{equation} \label{decoder6}
		\operatorname{mask}_i^t=\left\{\begin{array}{l}
			1, \text { if node } i \text { is masked at step } t \\
			0, \text { otherwise }.
		\end{array}\right.
	\end{equation} 
	
	Finally, the probability vector for vertex selection is generated by combining the compatibility vector between the vehicle and the vertices:
	\begin{equation} \label{decoder7}
		p^t=\operatorname{softmax}\left(\lambda^t+O \cdot \operatorname{mask} k^t\right),
	\end{equation}
	where $O$ is a large negative number, e.g., $O = -999999$, which means that the customer vertex cannot be visited.
	
	\subsubsection{Model training}	
		\begin{figure}[!t]
			\centering
			\includegraphics[width=3in]{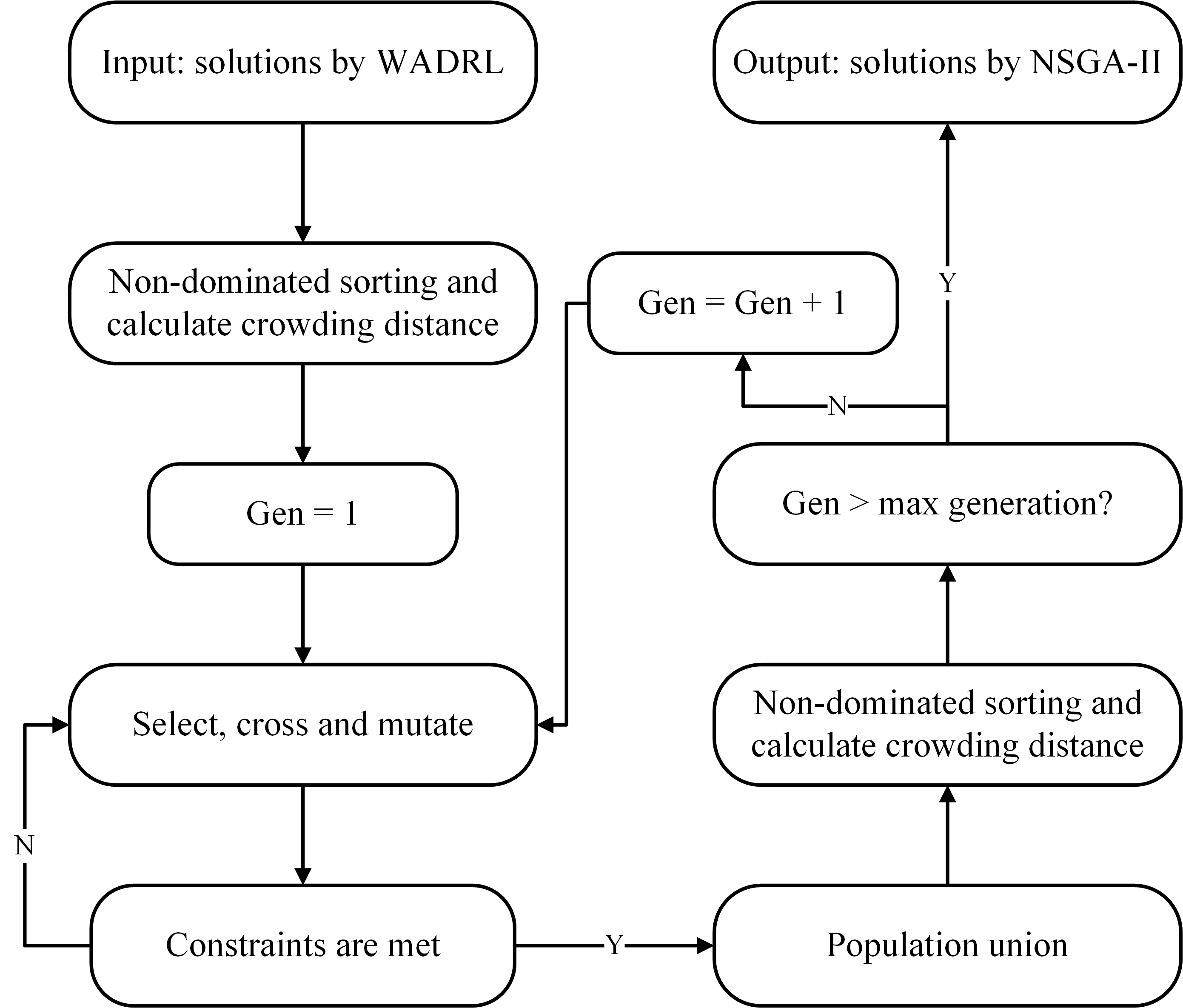}
			\caption{The process of combining WADRL and NSGA-II.}
			\label{fig_4}
		\end{figure}
	We employ a policy gradient method to train all parameters $\theta$ within the neural network. The algorithm comprises two networks, namely the policy network and the baseline network. Both the policy and baseline networks share the same network structure, with the only difference being that the policy network selects actions based on sampling from the probability vector, while the baseline network adopts a strategy where actions are selected based on the action possessing the highest probability, namely `greedy policy'. The gradient of the loss function is defined as follows \cite{alg2}
	\begin{equation}
		\nabla_\theta L(\theta)=E_{p_\theta\left(\pi^\theta\right)}\left[\left(R\left(\pi^\theta\right)-R\left(\pi^{B L}\right)\right) \nabla_\theta \log p_\theta\left(\pi^\theta\right)\right],
	\end{equation}
	where $R\left(\pi^\theta\right)$ and $R\left(\pi^{BL}\right)$ are utilized to denote the rewards accrued by the policy network and the baseline network, respectively. During each training batch, a t-test is employed to ascertain the statistical significance of the difference in performance between these two networks at a 95\% confidence level. Should this test yield a significant result, it prompts an update wherein the parameters of the policy network supersede those of the baseline network, thereby optimizing the learning process. The detailed procedural steps of this algorithm are given in Algorithm \ref{alg1}.

\subsection{Combine Weight-Aware Deep Reinforcement Learning with NSGA-II}

In the pursuit of solving the MOVRPTW, the application of WADRL may reveal promising results. However, several limitations still exist: \ding{172} The neural network sometimes struggles to iterate optimally, falling into a local optimal solution; \ding{173} Its performance may exhibit instability under specific objective weight combinations; \ding{174} Most of the obtained solutions may be dominated by other solutions. 

To address these limitations, we propose the approach that combines WADRL with NSGA-II. Our approach involves utilizing WADRL to generate initial solutions for the MOVRPTW. These initial solutions serve as the foundation for subsequent optimization through NSGA-II. WADRL ensures that the initial solutions are of high quality, meet all constraints, and have a relatively uniform distribution due to the applied decomposition strategy.

This combined approach is designed to improve the overall optimization process, enhance solution quality, and mitigate the inherent limitations of WADRL. The integration of NSGA-II complements the strengths of WADRL, providing a more robust and effective solution strategy for MOVRPTW. The process of NSGA-II optimizing the solution generated by WADRL is shown in Fig. \ref{fig_4}.

\begin{table*}\label{table_2}
	\centering
	\caption{Average Objective Functions Values and Running Time of Generating Initial Solutions by WADRL and NSGA-II. Instances of 20-, 50- and 100- customer Are Listed.}
	\begin{tabular}{c c c c c c c c c c}
		\toprule[1.5pt]
		& \multicolumn{3}{c}{20-customer} &  \multicolumn{3}{c}{50-customer}&  \multicolumn{3}{c}{100-customer} \\
		\toprule[1pt]
		& $f_1$ & $f_2$ & Time/s & $f_1$ & $f_2$ &Time/s& $f_1$ & $f_2$ & Time/s \\
		\toprule[1pt]
		
		WADRL& 2812.98&0.7464&2.11&5481.24&0.9072&4.59&13953.19&0.9331&7.13 \\
		
		NSGA-II& 3705.36& 0.5452& 2.07 &9640.94 &0.6013 &9.12 &17961.76 &0.5616 &35.58 \\
		
		\toprule[1.5pt]
	\end{tabular}
\end{table*}

\begin{figure*}[!t]
	\centering
	\subfloat[20-customer instance] {\includegraphics[width=2.3in]{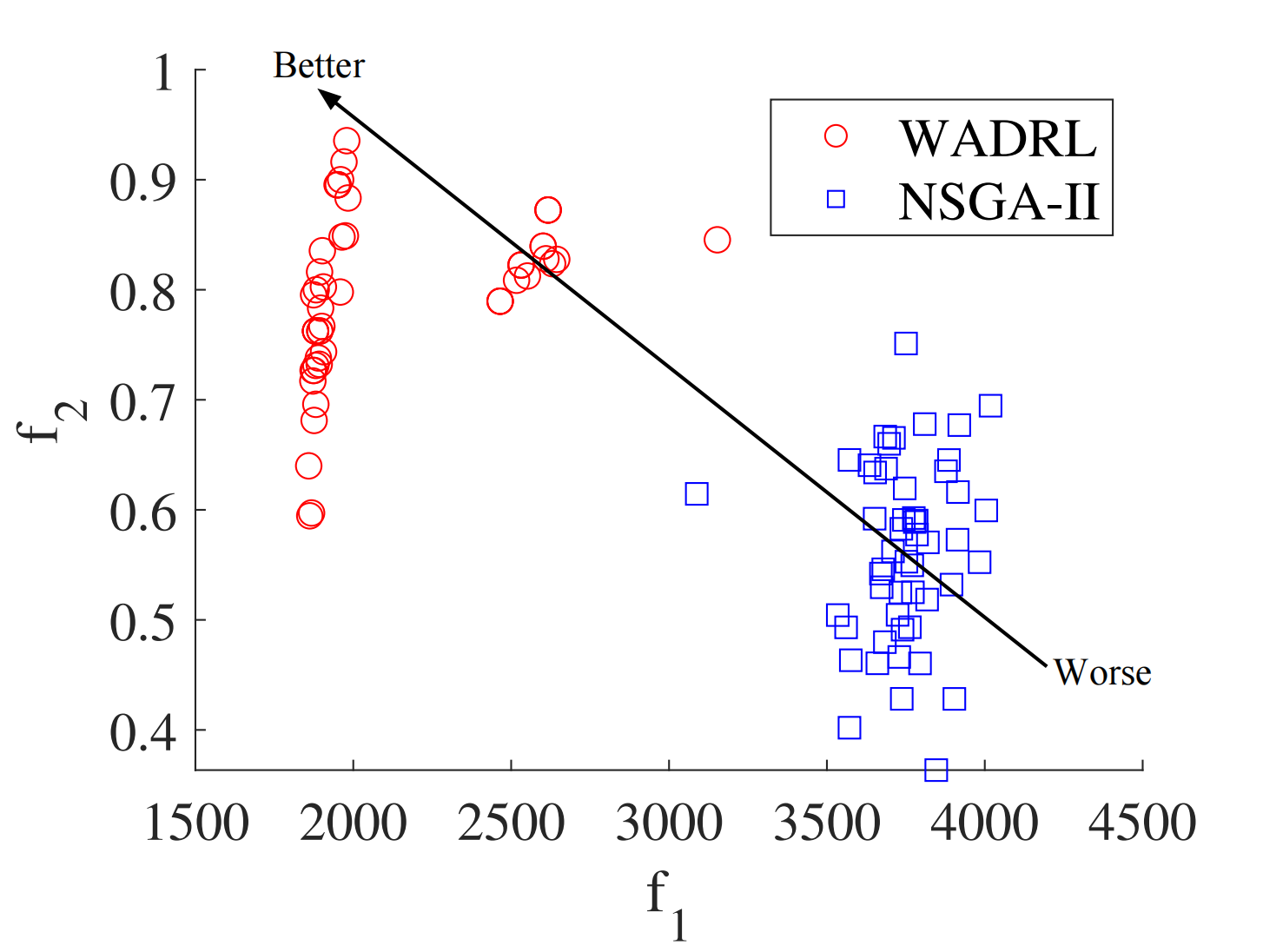}%
		\label{init_20city}}
	\hfil
	\subfloat[50-customer instance] {\includegraphics[width=2.3in]{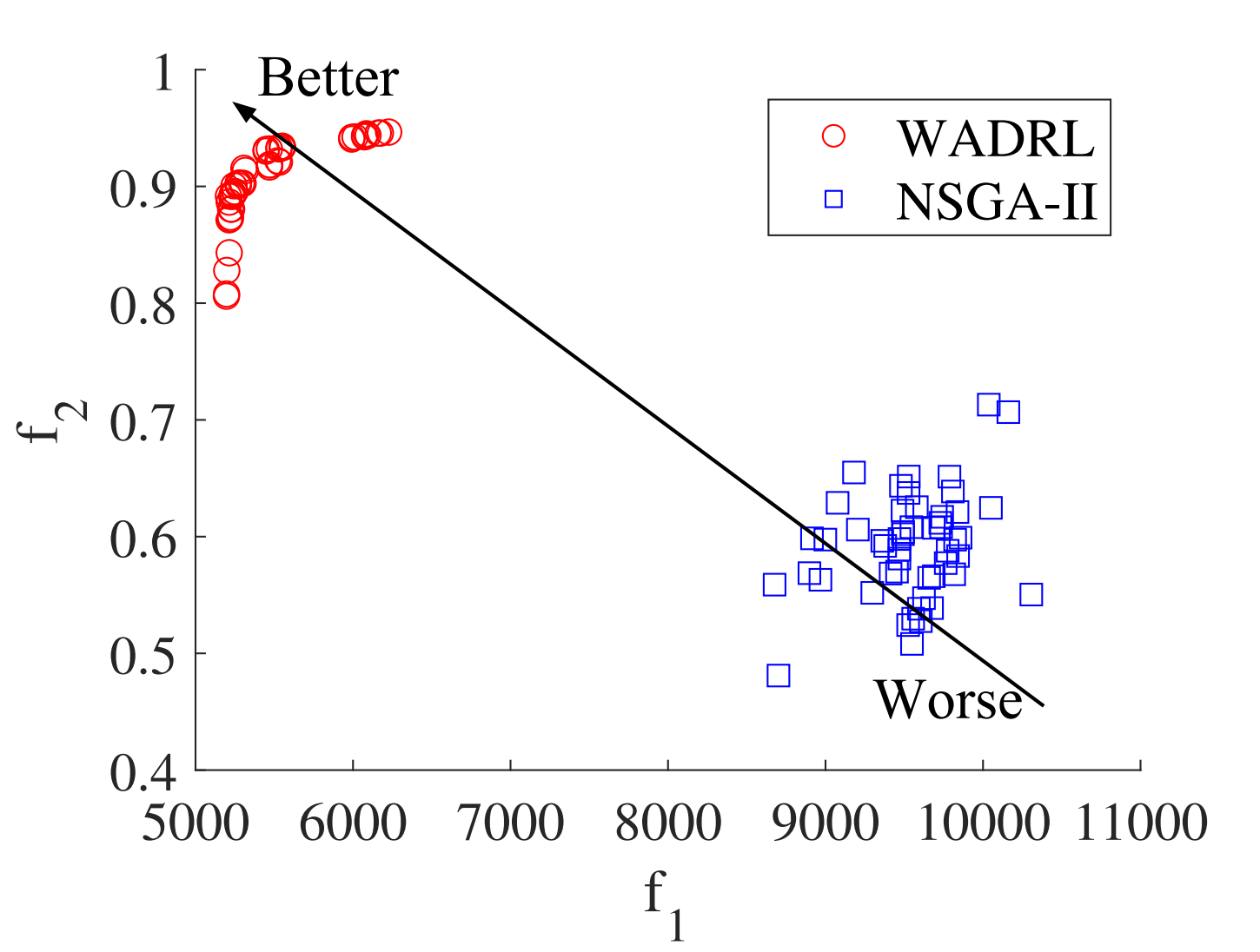}%
		\label{init_50city}}
	\hfil
	\subfloat[100-customer instance] {\includegraphics[width=2.3in]{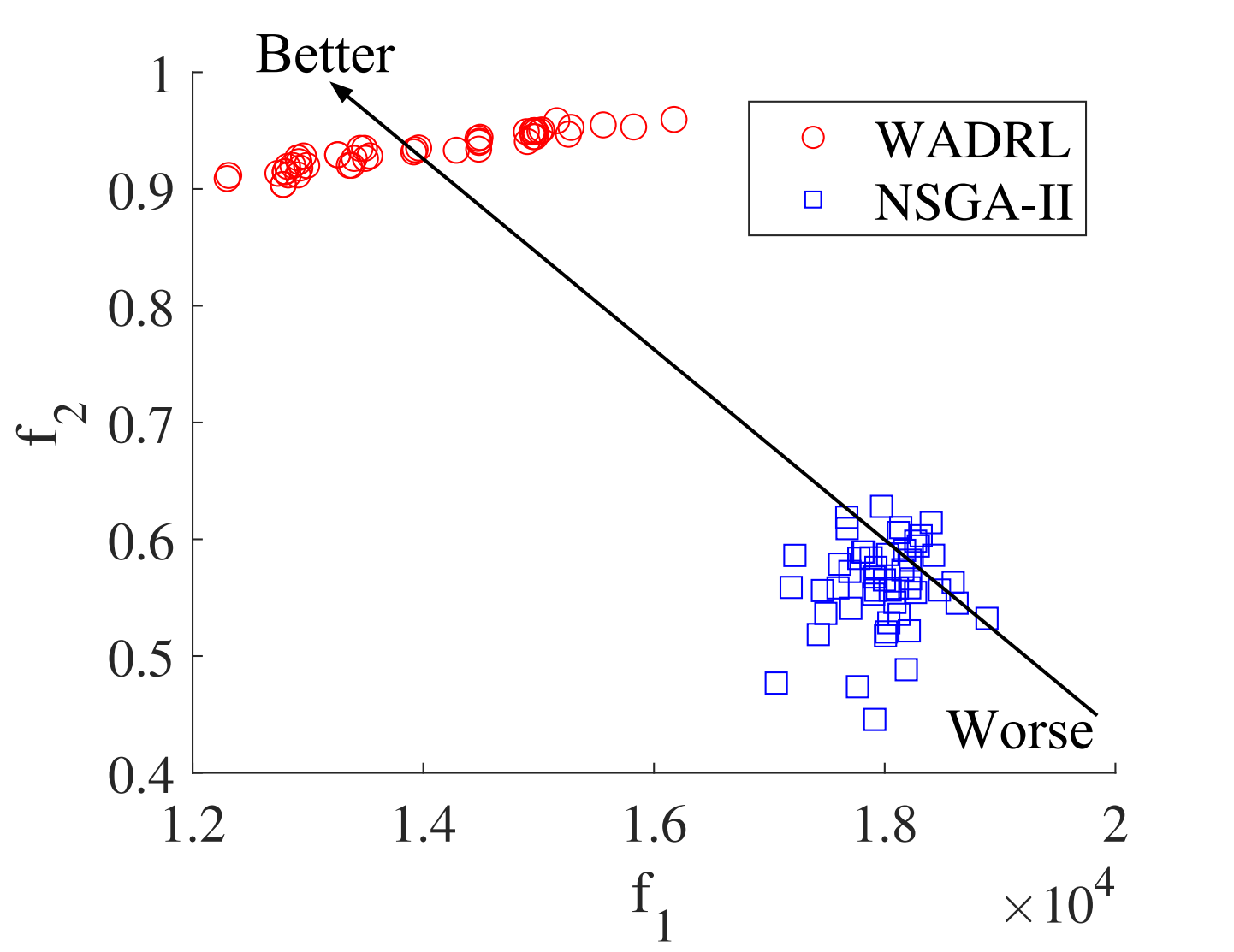}%
		\label{init_100city}}
	\caption{Initial solutions generated by WADRL and NSGA-II in the instances of 20-, 50- and 100-customer.}
	\label{init}
\end{figure*}

\begin{table*}\label{table_3}
	\centering
	\caption{Average Objective Functions Values of the PFs by WADRL+NSGA-II, NSGA-II, WADRL and DRL. Instances of 20-, 50- and 100- customer Are Listed. The Running Time Are Also Listed.}
	\begin{tabular}{c c c c c c c c c c}
		\toprule[1.5pt]
		& \multicolumn{3}{c}{20-customer} &  \multicolumn{3}{c}{50-customer}&  \multicolumn{3}{c}{100-customer} \\
		\toprule[1pt]
		& $f_1$ & $f_2$ & Time/s & $f_1$ & $f_2$ &Time/s& $f_1$ & $f_2$ & Time/s \\
		\toprule[1pt]
		
		NSGAII-200& 2913.21 & 0.7377& 9.18& 7825.94& 0.7977& 60.29& 14993.88& 0.6132 & 169.51 \\
		
		NSGAII-500& 2526.57&0.7421& 18.69& 6536.26& 0.7895& 72.96& 14761.331& 0.7754 & 216.09 \\
		
		NSGAII-1000& 1929.04&0.8584 & 35.04& 5813.88& 0.8327& 112.35& 13581.44& 0.7831& 620.48\\
		
		NSGAII-2000& 1916.32&0.8511 & 68.41&5323.13 & 0.8237& 245.65& 13139.78& 0.8780 & 1126.98\\
		
		DRL & 1923.99&0.78 & 3.15&5424.69 &0.91 & 5.41& 14026.93& 0.9293& 8.56\\
		
		WADRL & 1916.66&0.8284 & 2.11&5584.11 &0.9122 & 4.59& 13857.09& 0.9346& 7.13\\
		
		WADRL+NSGAII-500 & 1911.56&0.8203 & 15.55&5339.38 &0.9126 & 61.22& 12085.43& 0.9284& 193.41\\
		\toprule[1.5pt]
	\end{tabular}
\end{table*}

\begin{figure}[!t]
	\centering
	\includegraphics[width=3in]{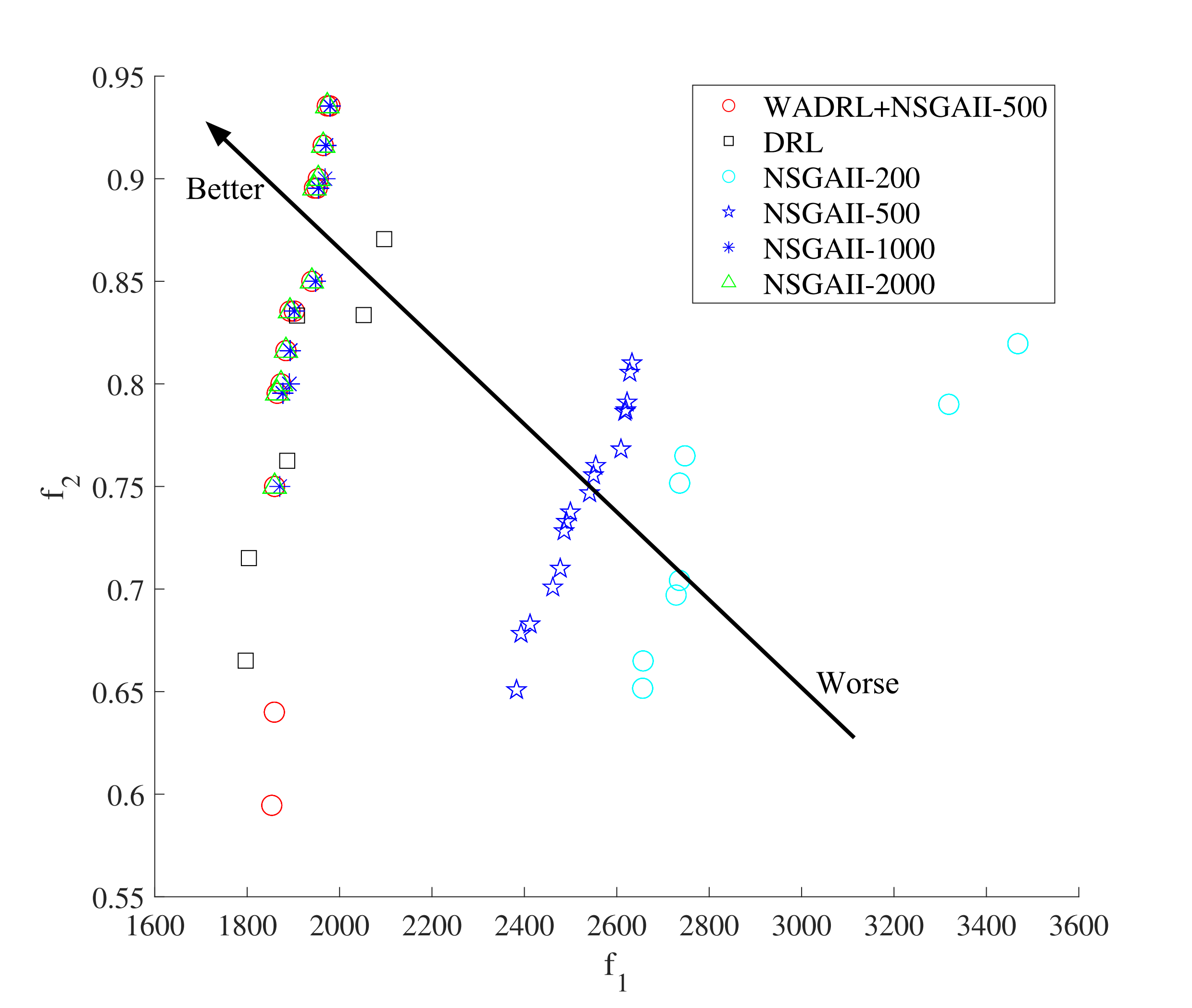}
	\caption{20-customer MOVRPTW problem instance: The Pareto front obtained using our method in comparison with DRL and NSGA-II. 200, 500, 1000, 2000 iterations are applied respectively.}
	\label{fig_6}
\end{figure}

\begin{figure}[!t]
	\centering
	\includegraphics[width=3in]{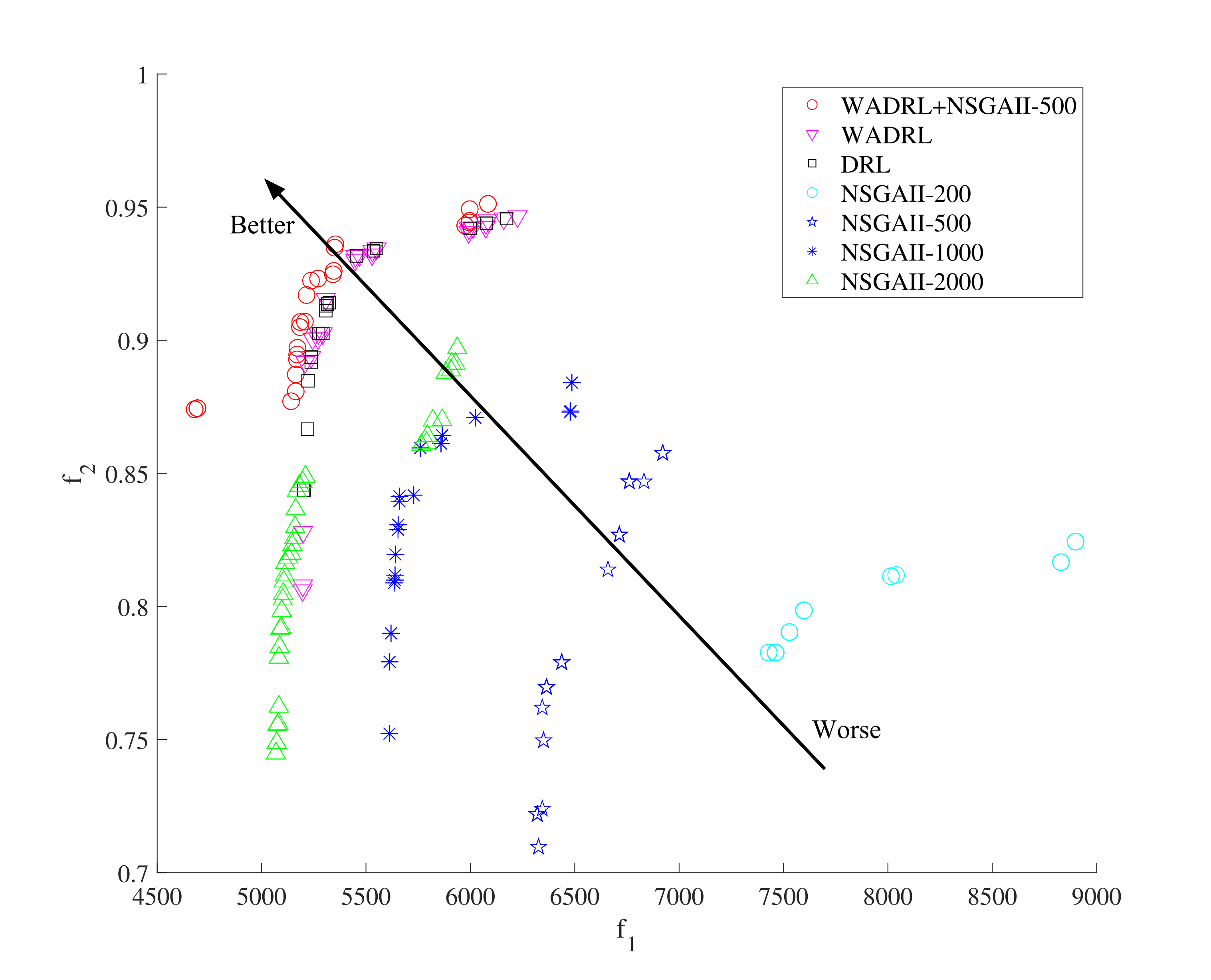}
	\caption{50-customer MOVRPTW problem instance: The Pareto front obtained using our method in comparison with DRL, WADRL and NSGA-II. 200, 500, 1000, 2000 iterations are applied respectively.}
	\label{fig_7}
\end{figure}

\begin{figure}[!t]
	\centering
	\includegraphics[width=3in]{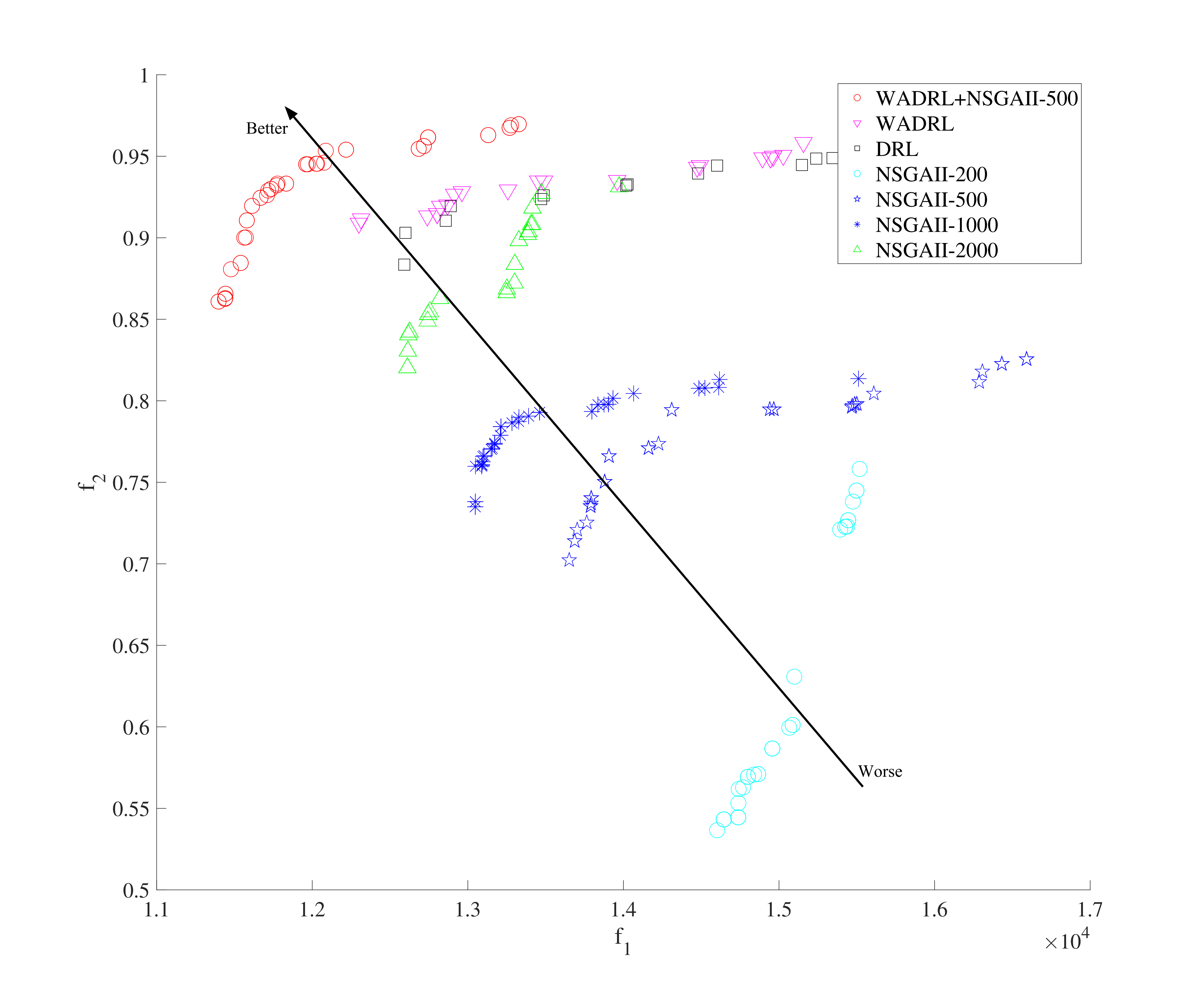}
	\caption{100-customer MOVRPTW problem instance: The Pareto front obtained using our method in comparison with DRL, WADRL and NSGA-II. 200, 500, 1000, 2000 iterations are applied respectively.}
	\label{fig_8}
\end{figure}

\begin{figure*}[!t] 
	\centering
	\subfloat[20-customer instance]{\includegraphics[width=2.3in]{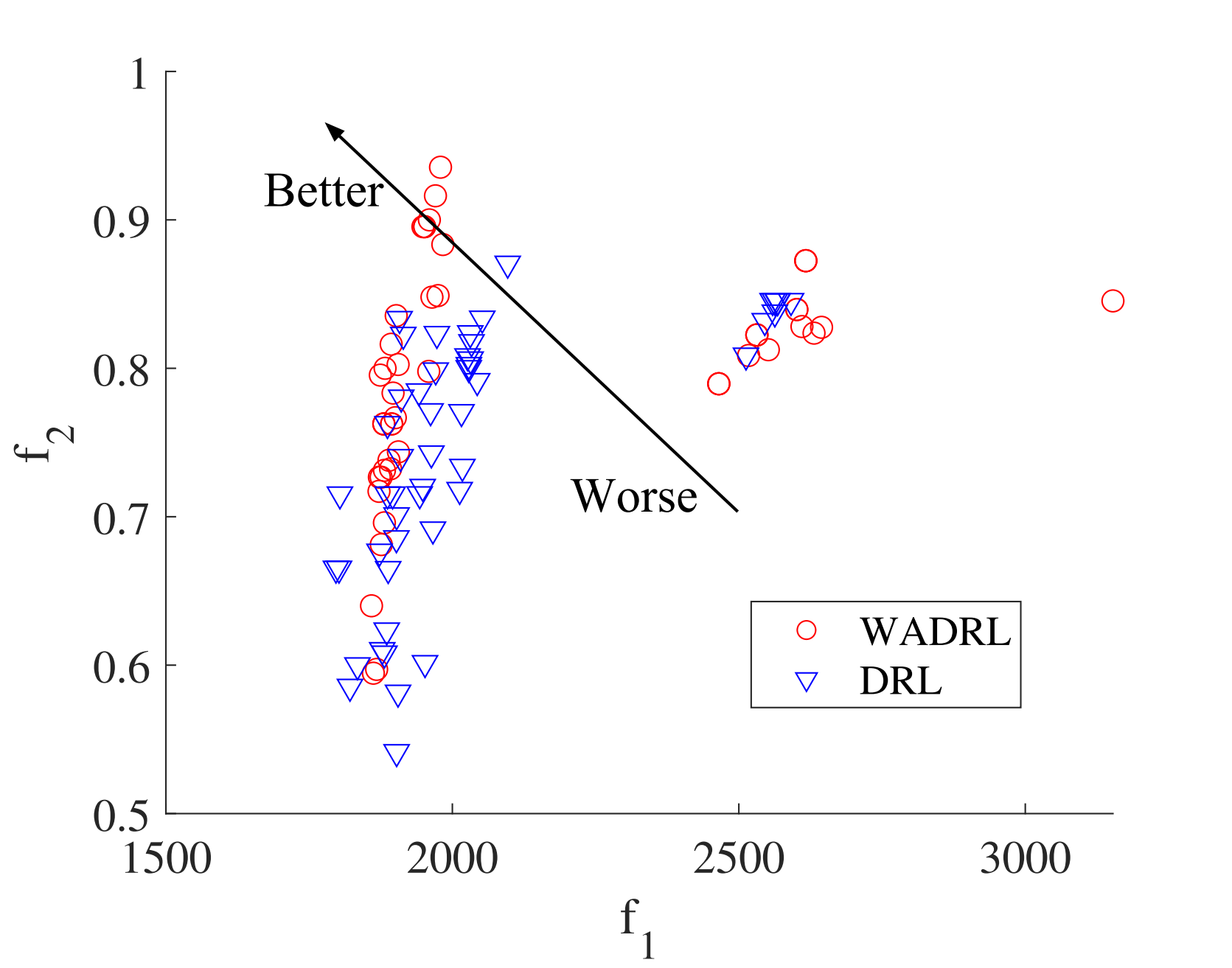}
		\label{fig91}}%
	\hfil
	\subfloat[50-customer instance]{\includegraphics[width=2.3in]{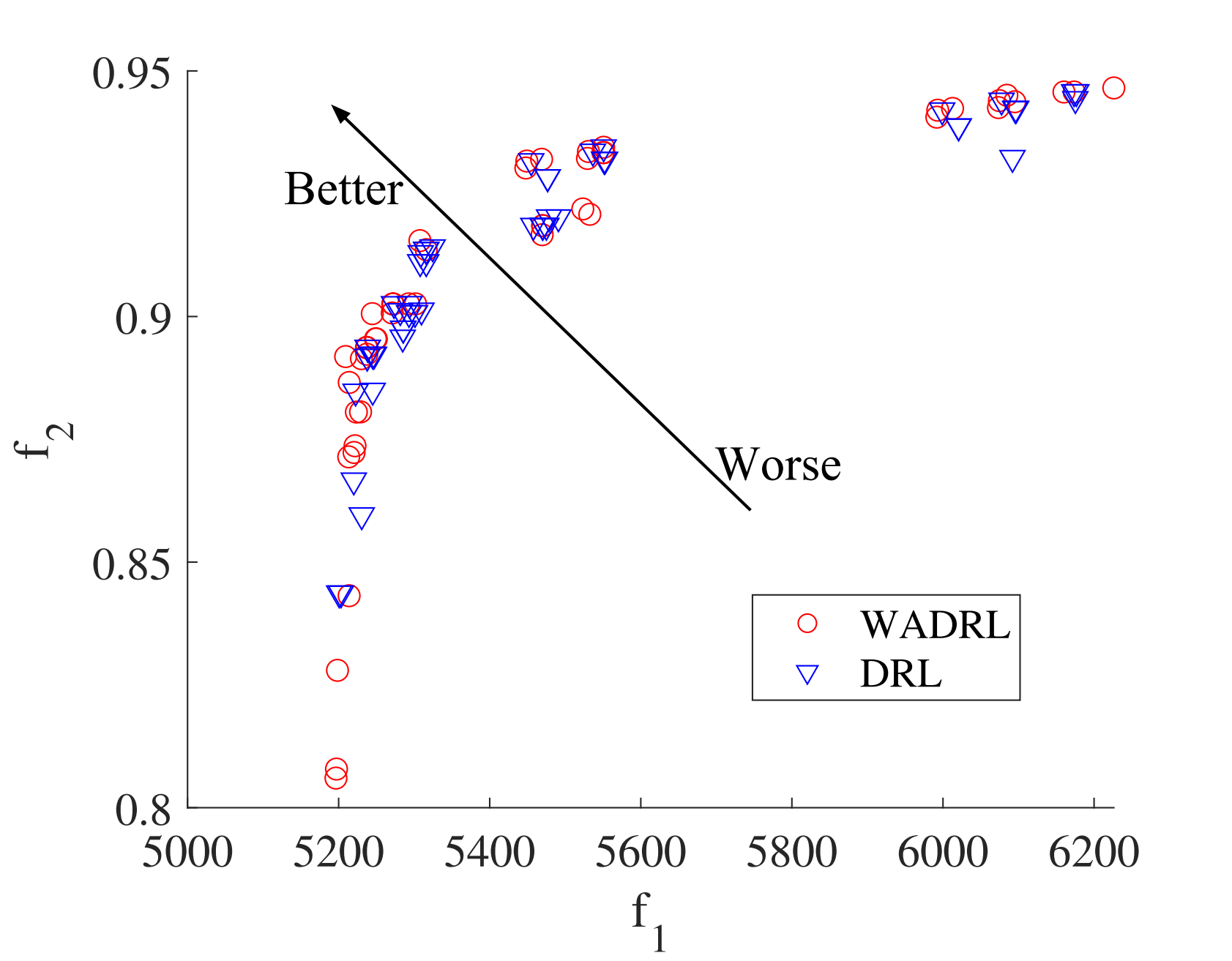}\label{fig92}}%
	\hfil
	\subfloat[100-customer instance] {\includegraphics[width=2.3in]{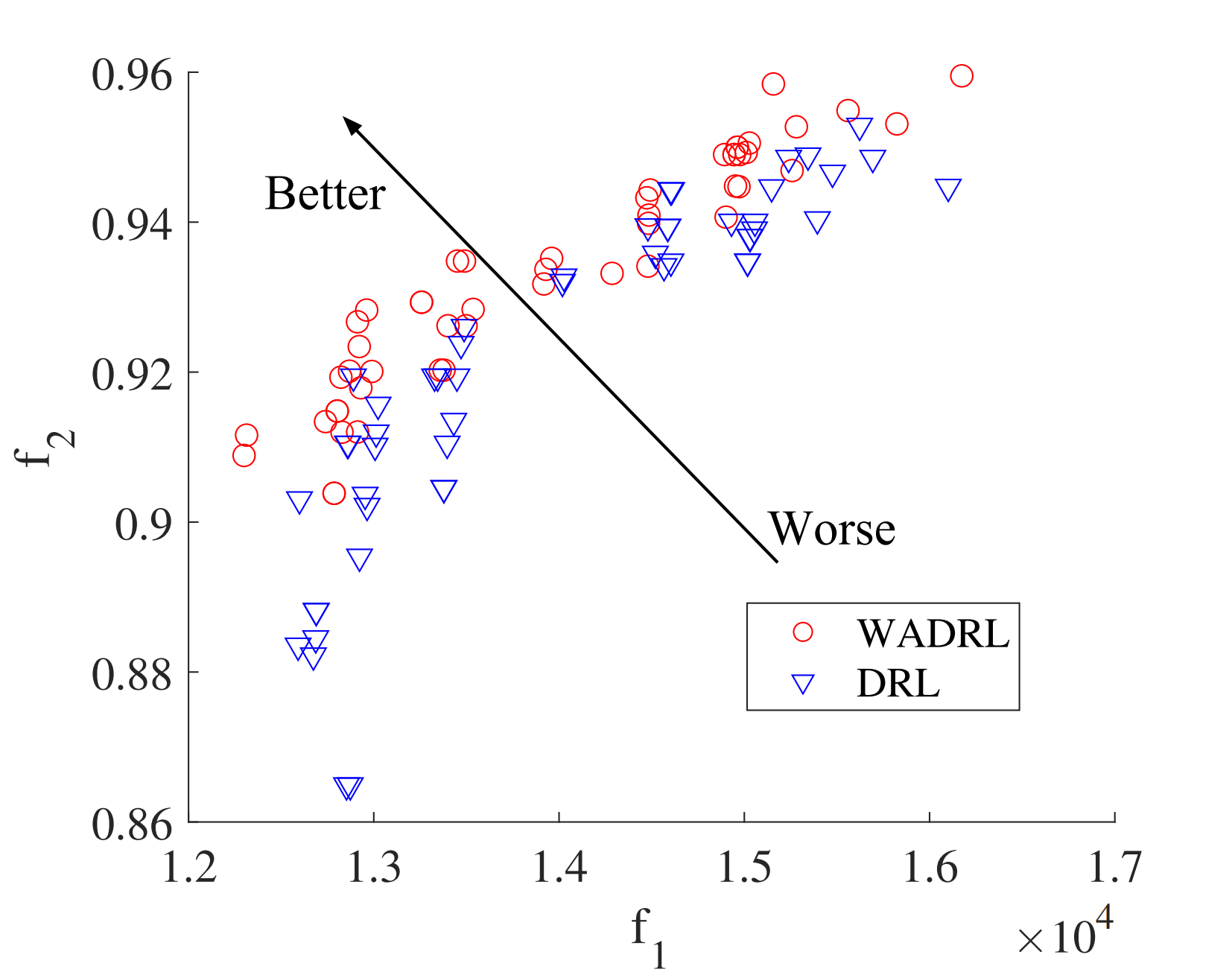}\label{fig93}}%
	\caption{Solutions generated by WADRL and DRL in the instances of 20-, 50- and 100-customer.}
	\label{fig_9}
\end{figure*}

\section{Results and Analysis}

To demonstrate the advantages of our proposed approach that combines weight-aware deep reinforcement learning (WADRL) and non-dominated sorting genetic algorithm-II (NSGA-II), we apply it to the Solomon dataset \cite{solomon2008}. Furthermore, we conduct comparative studies on datasets with varying numbers of customers, comparing our method with state-of-the-art approaches, including traditional deep reinforcement learning (DRL) and NSGA-II.

\subsection{Experimental Setup}

\subsubsection{Training sets}

The model is trained on instances with 20-, 50- and 100-customer vertices and a single depot vertex, respectively. In the training dataset, the demand for customer vertices is uniformly sampled from the range $[1,40]$. For the instances with 20-, 50-, and 100-customer instances, the coordinates of the vertices are generated from a uniform distribution within the range $[0, 100]$. We randomly generate soft time windows with intervals greater than 30 within the range $[0, 240]$.

\subsubsection{Test set}

For the 20-customer instance, we use the information of the first 20 customers in RC101; For the 50-customer instance, we use the information of the first 50 customers in RC101; For the 100-customer instance, we use RC102 \cite{solomon}.

\subsubsection{Parameter settings}

In MOVRPTW, the travel cost of vehicle $k$ per unit distance and fixed cost generated by using vehicle, denoted as $k$ $\left(c_k^1, c_k^2\right)$, are set to be fixed, as described in $(2.0,400)$ \cite{movrptw1}. The allowable maximum violation of time window parameters, denoted as $\left(\zeta_i^e, \zeta_i^l\right)$, are set as $(0.25,0.25)$.

In each iteration, a set of weight combinations is extracted according to the Dirichlet distribution. We use the Adam optimizer with a learning rate of $10^{-4}$ with decay of $10^{-6}$ \cite{weight1}. In total, we train the model for 300 epochs and the batch size is set to 64.

DRL and WADRL use weight combinations with an interval of 0.02 to decompose MOVRPTW. Specifically, these weight combinations are defined as [[1.00, 0.00], [0.98, 0.02], ... , [0.00, 1.00]], resulting in a total of 51 subproblems. Furthermore, we configure the population size of the NSGA-II algorithm to also be 51.

\subsection{Comparison of Initial Solutions}
We first compare the performance of our method and the original NSGA-II in generating initial solutions, as shown in Fig. \ref{init}. The results indicate that our method consistently outperforms the traditional NSGA-II algorithm in terms of the quality of initial solutions, across instances with 20-, 50-, 100-customer. Furthermore, the average execution time together with the average objective function value of the initial solution are presented in TABLE III.

As shown in Fig. \ref{init}, all initial solutions generated by the WADRL-based NSGA-II outperform the initial solutions generated by the original NSGA-II using random policies, both in terms of objective functions and time efficiency. This improvement is particularly pronounced as the number of customers increases, with our method exhibiting more significant enhancements over the traditional NSGA-II approach. For example, in the case of the 100-customer instance, our method, on average, reduces the cost by 4008.6, increases customer satisfaction by 0.3715, and decreases the time required for generating initial solutions by 28.45 seconds. Furthermore, we observe that within the solutions generated by WADRL, many of them are dominated by other solutions, rendering them non-meaningful in MOVRPTW. As a result, it is reasonable to use the NSGA-II algorithm to optimize these solutions generated by WADRL.

\subsection{Experimental Results}

We implement NSGA-II algorithms with 200, 500, 1000, and 2000 iterations, as well as NSGA-II with 500 iterations using the solution of WADRL as the initial solution. We compare these results with DRL and WADRL, and the average objective function values and running times are presented in TABLE III.

Fig. \ref{fig_6}, Fig. \ref{fig_7} and Fig. \ref{fig_8} depict the experimental results on instances of 20-, 50-, and 100-customer, respectively. Our approach (WADRL+NSGA-II) consistently achieves optimal results on all instances, particularly as the problem scale increases. WADRL itself has achieved relatively competitive results, and when combined with the NSGA-II algorithm, WADRL's solutions exhibit higher quality and extended coverage on the Pareto front.

In the 20-customer instance, our method, NSGAII-1000 and NSGAII-2000 show almost the same performance, but our method still finds solutions which have a lower travel cost. Not to mention that in the 100-customer instance, the advantage of our method is overwhelming, using less than 200 seconds to obtain effects that the NSGA-II algorithm cannot obtain in 1000 seconds.

In terms of the average objective function values, as shown in TABLE III, the WADRL+NSGA-II algorithm outperforms other benchmark algorithms under all instances. Meanwhile, the runtime of WADRL is significantly lower than that of the traditional DRL algorithms and all NSGA-II algorithms.

\subsection{Effectiveness of Weight-aware Strategy}
\begin{table}\label{table:WA}
	\centering
	\caption{The Training Time of WADRL and DRL. Instances of 20-, 50- and 100-customer Are Listed.}
	\begin{tabular}{c c c c}
		\toprule[1.5pt]
		&20-customer &  50-customer&  100-customer \\
		\toprule[1pt]
		WADRL& 24min & 48min & 1h43min \\
		
		DRL& 1h54min & 5h17min & 17h32min  \\
		
		\toprule[1.5pt]
	\end{tabular}
\end{table}
In this section, the weight-aware strategy in deep reinforcement learning is experimentally validated. Fig. \ref{fig_9} displays the results of all solutions generated by WADRL and DRL on instances on 20-, 50-, and 100-customer instances. TABLE IV presents the average training times. The results indicate that WADRL achieves superior results while utilizing only about 1/10 of the training time compared of the traditional DRL.

In the 20-customer instance, as shown in Fig. \ref{fig91}, the results of DRL are relatively unstable, and the results of WADRL are relatively stable. In the 50-customer instance, as shown in Fig. \ref{fig92}, the results of DRL and WADRL are almost the same, but WADRL still finds some solutions with lower travel cost. In the 100-customer instance, as shown in Fig. \ref{fig93}, the results of WADRL are significantly better than those of DRL.

\section{Conclusion}
This paper proposed a weight-aware deep reinforcement learning (WADRL) framework combined with non-dominated sorting algorithm-II (NSGA-II) to solve the multiobjective vehicle routing problem with time windows (MOVRPTW). Firstly, a comprehensive MOVRPTW model was proposed, which takes into account both travel cost and customer satisfaction. Subsequently, a WADRL framework was introduced, where the weights of objective functions were integrated into the state of deep reinforcement learning (DRL), enabling a single DRL model to solve the entire multiobjective optimization problem. An innovative DRL framework was introduced, which is centered around a transformer-based policy network. The architecture of this policy network was carefully designed, comprising three integral components: the encoder module, which is responsible for embedding customer information, the weight embedding module, which plays a critical role in capturing and encoding the weights of objective functions, and the decoder module, tasked with generating executable actions based on the contextual information. Furthermore, considering the limitations of DRL, we employed the NSGA-II algorithm to optimize the solutions generated by WADRL. Experimental results demonstrated the promising performance of our method across all MOVRPTW instances. Specifically, our method produced Pareto fronts with better coverage and quality. Finally, the weight-aware strategy reduces the training time of DRL greatly and achieves superior performance.

Regarding future research, in the current use of WADRL to solve MOVRPTW, a single DRL model can solve the entire MOVRPTW for the same scale of customers. However, for different MOVRPTW scales, multiple models still need to be trained. We will investigate how to train a single DRL model to handle MOVRPTW instances of varying scales.

\bibliography{TITS_ref}

\begin{IEEEbiography}[{\includegraphics[width=0.8in,height=1in,clip,keepaspectratio]{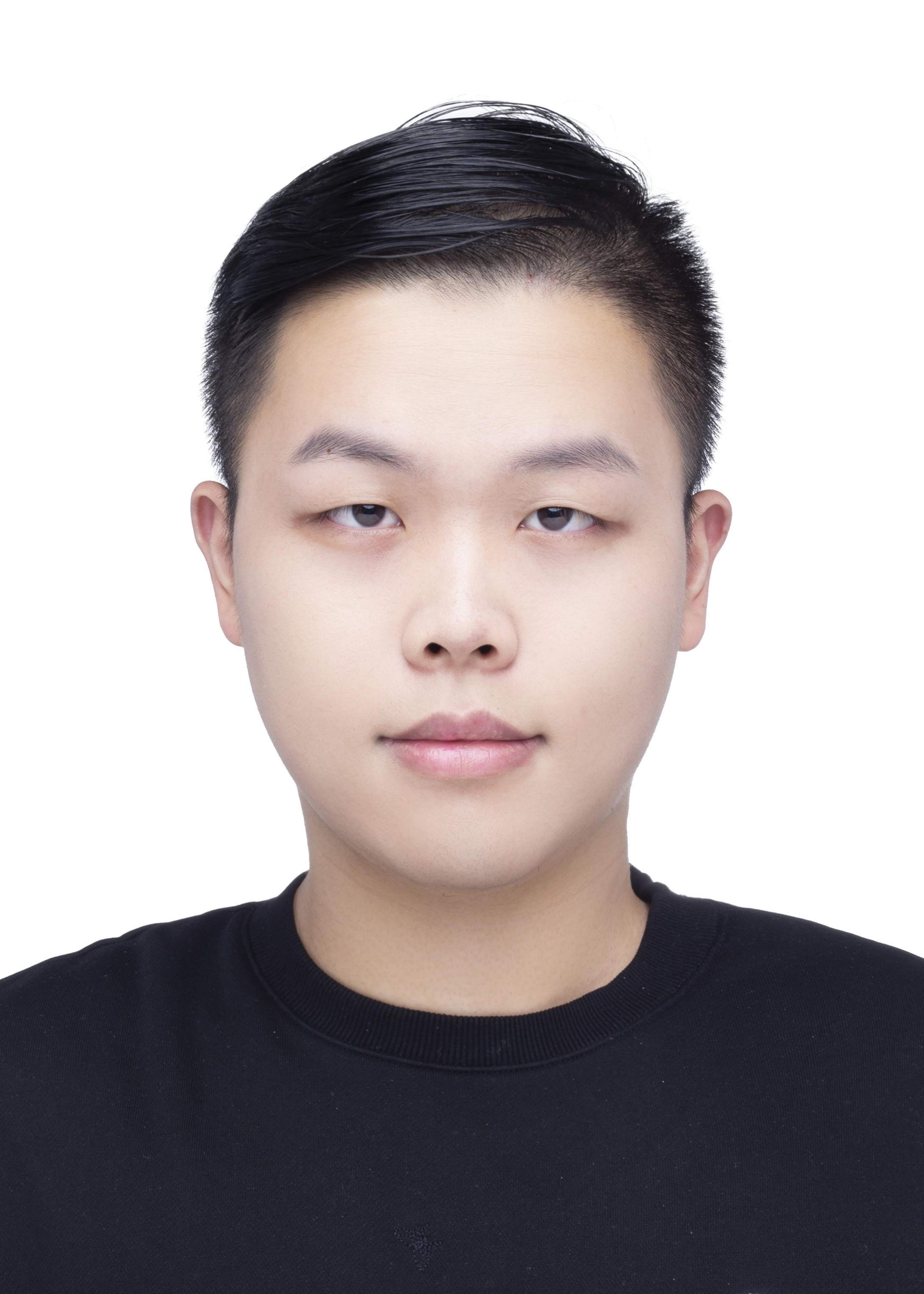}}]{Rixin Wu} was born in Taizhou, Jiangsu, China in 2000, and obtained an B.E. degree from Nanjing University of Chinese Medicine in 2022 and is currently pursuing a M.E degree at Nanjing University of Aeronautics and Astronautics. His research focuses on deep reinforcement learning and multiobjective optimization problems.
\end{IEEEbiography}

\begin{IEEEbiography}[{\includegraphics[width=0.8in,height=1in,clip,keepaspectratio]{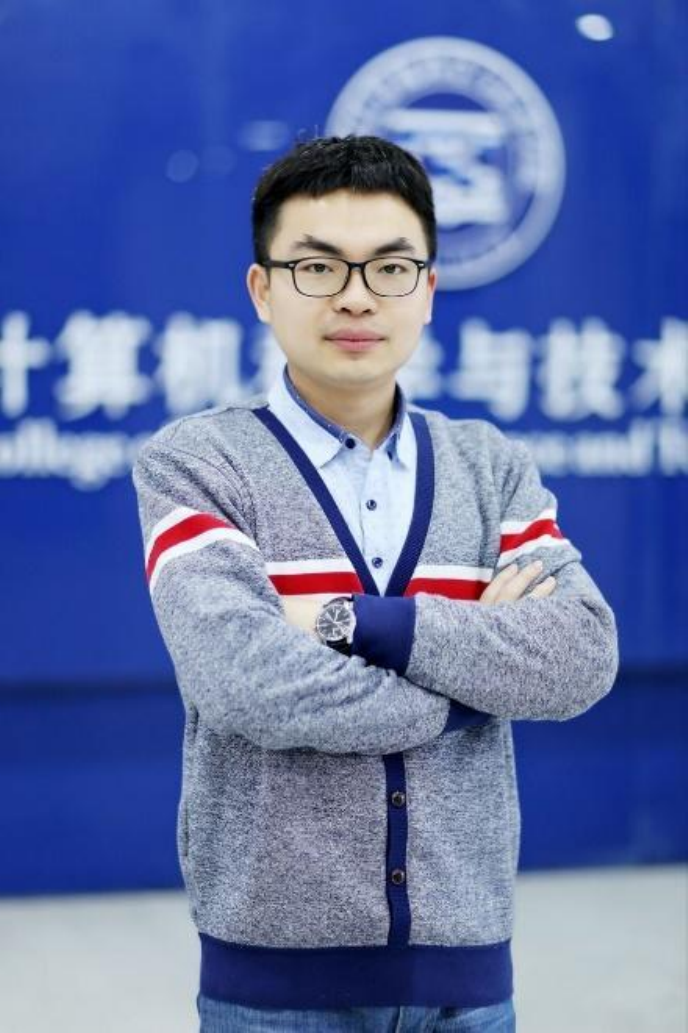}}]{Ran Wang}
	(Member, IEEE) is an Associate professor and Doctoral Supervisor with the College of Computer Science and Technology, Nanjing University of Aeronautics and Astronautics, Nanjing, China. He received his B.E. in Electronic and Information Engineering from Honors School, Harbin Institute of Technology, China in July 2011 and Ph.D. in Computer Science and Engineering from Nanyang Technological University, Singapore in April 2016. He has authored or co-authored over 50 papers in top-tier journals and conferences. He received the Nanyang Engineering Doctoral Scholarship (NEDS) Award in Singapore and the innovative and entrepreneurial Ph.D. Award of Jiangsu Province, China in 2011 and 2017, respectively. He is the recipient of the Second Prize for Scientific and Technological Progress awarded by the China Institute of Communications and he is the ChangKong Scholar of NUAA. His current research interests include telecommunication networking and cloud computing.
\end{IEEEbiography}

\begin{IEEEbiography}[{\includegraphics[width=0.8in,height=1in,clip,keepaspectratio]{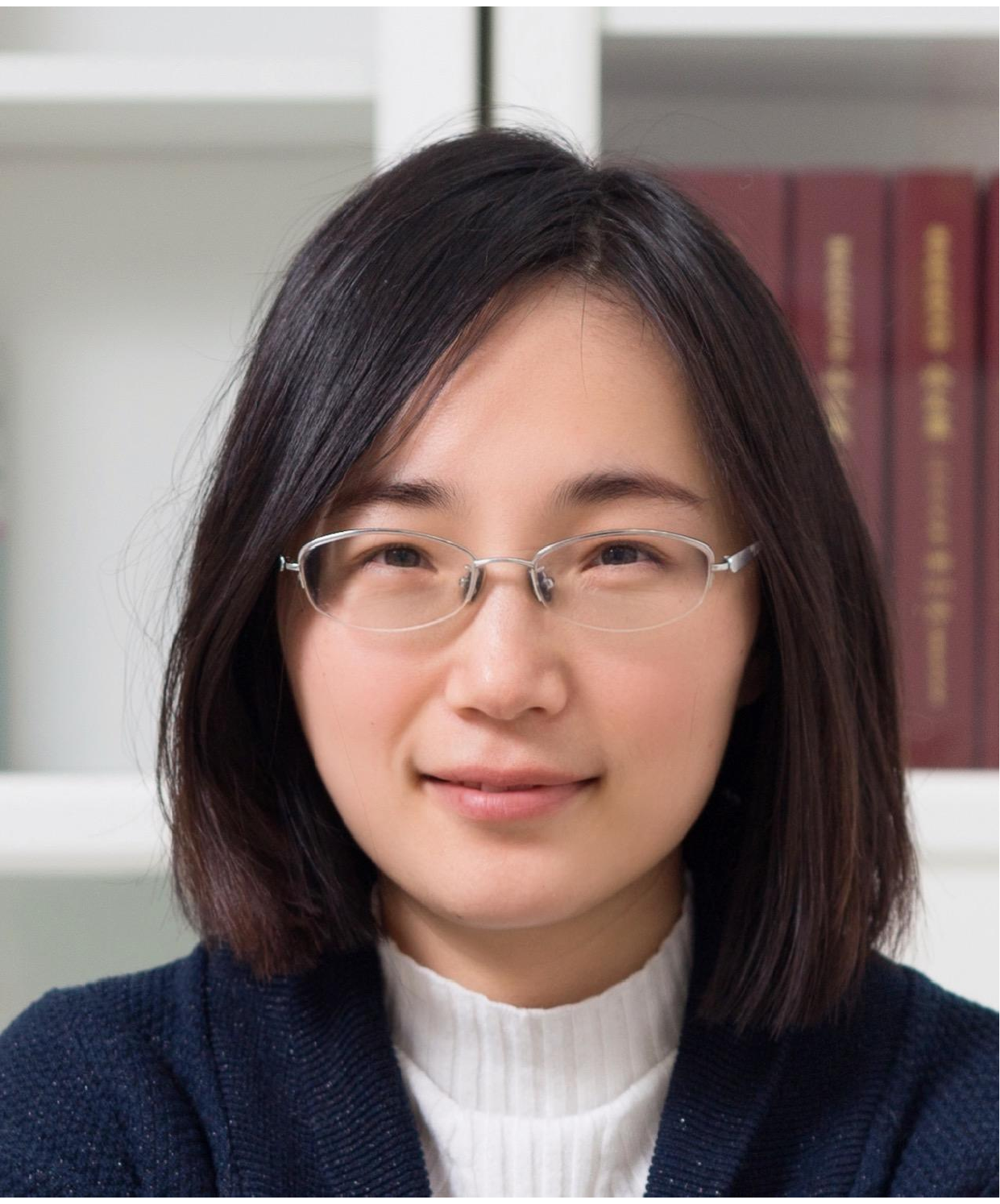}}]{Jie Hao} 
	(Member IEEE) received her BS degree from Beijing University of Posts and Telecommunications, China, in 2007, and the Ph.D. degree from University of Chinese Academy of Sciences, China, in 2014. From 2014 to 2015, she has worked as post-doctoral research fellow in the School of Computer Engineering, Nanyang Technological University, Singapore. She is currently an Associate Professor at College of Computer Science and Technology, Nanjing University of Aeronautics and Astronautics, China. Her research interests are wireless sensing, wireless communication and etc.
\end{IEEEbiography}

\begin{IEEEbiography}[{\includegraphics[width=0.8in,height=1in,clip,keepaspectratio]{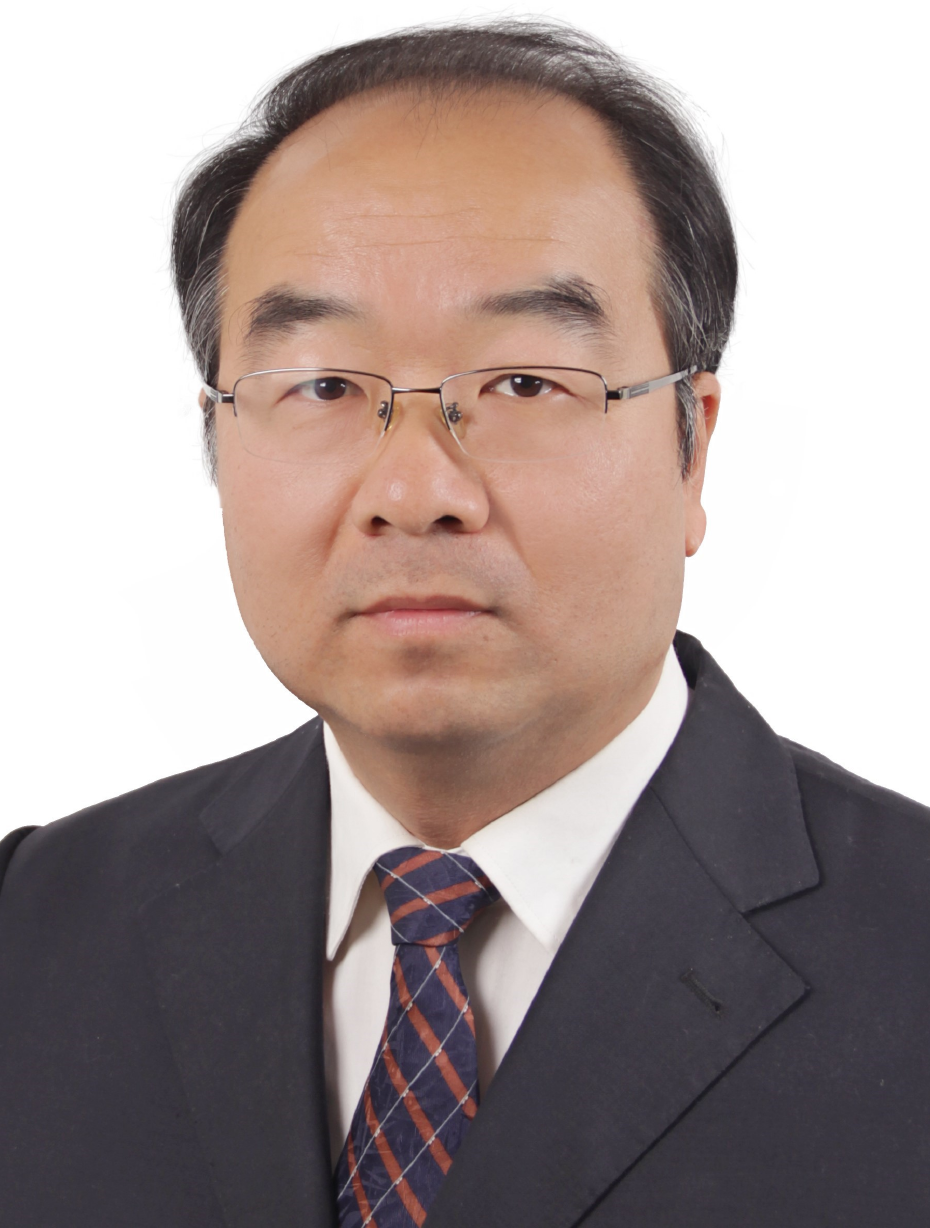}}]{Qiang Wu}
	(Member, IEEE) is currently a Professor with the College of Computer Science and Technology, Nanjing University of Aeronautics and Astronautics, Nanjing, China. Before that, he worked as a general engineer of ZTE Central Research Institute, and received his Ph.D. degree from the College of Computer Science and Technology, Zhejiang University, Hangzhou, China. Prof. Wu is a Fellow of the China Institute of Communications, an Executive Member of Service Computing Technical Committee of CCF, and a Member of the Technical Committee of National Engineering Laboratory for Mobile Internet System and Application Security. His main research fields include future networks, industrial Internet, cyber security, and space-earth integration networks. The Chinese government honored him with the Second-Class National Science and Technology Progress Award in $2009$ and the Second-Class National Technology Innovation Award in 2014. He has more than $100$ authorized patents, of which more than $20$ have corresponding relations with international standards.
\end{IEEEbiography}

\begin{IEEEbiography}[{\includegraphics[width=0.8in,height=1in,clip,keepaspectratio]{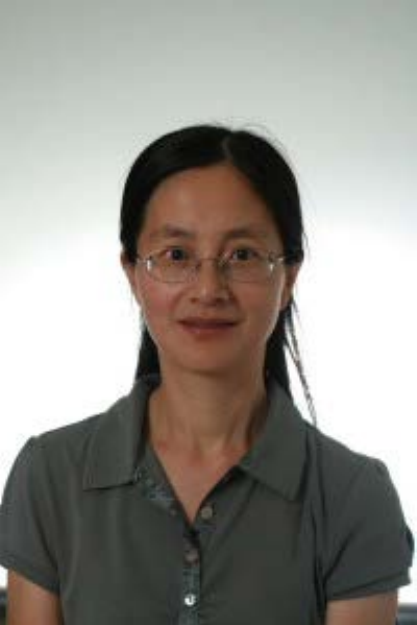}}]{Ping Wang}
	(Fellow, IEEE) is a Professor at the Department of Electrical Engineering and Computer Science, York University, and a Tier 2 York Research Chair. Prior to that, she was with Nanyang Technological University, Singapore, from 2008 to 2018. Her research interests are mainly in radio resource allocation, network design, performance analysis and optimization for wireless communication networks, mobile cloud computing and the Internet of Things. Her scholarly works have been widely disseminated through top-ranked IEEE journals/conferences and received four Best Paper Awards from IEEE prestigious conferences. Her publications received 28,900+ citations with H-index of 77 (Google Scholar). She is a Fellow of IEEE and a Distinguished Lecturer of the IEEE Vehicular Technology Society.
\end{IEEEbiography}

\begin{IEEEbiography}[{\includegraphics[width=0.8in,height=1in,clip,keepaspectratio]{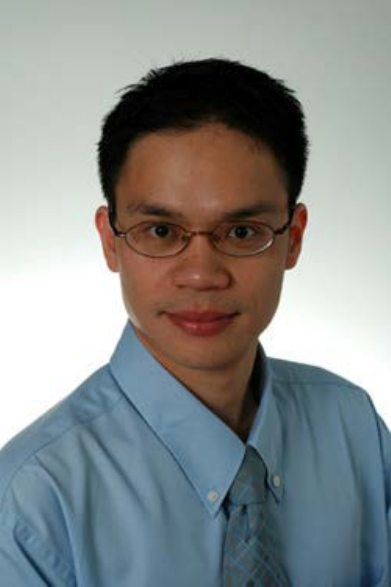}}]{Dusit Niyato}
	(Fellow, IEEE) received a Ph.D. degree in Electrical and Computer Engineering from the University of Manitoba, Canada, in 2008. He is currently a Professor in the School of Computer Science and Engineering at Nanyang Technological University, Singapore. He has published more than 400 technical papers in the area of wireless and mobile computing. He has won the Best Young Researcher Award of the Asia/Pacific chapter of the IEEE Communications Society and the 2011 IEEE Communications Society Fred W. Ellersick Prize Paper Award. Currently, he is serving as a senior editor of IEEE Wireless Communications Letters, an area editor of IEEE Transactions on Wireless Communications, an area editor of IEEE Communications Surveys and Tutorials, an editor of IEEE Transactions on Communications, and an associate editor of IEEE Transactions on Mobile Computing, IEEE Transactions on Vehicular Technology, and IEEE Transactions on Cognitive Communications and Networking. He was a Distinguished Lecturer of the IEEE Communications Society in 2016-2017. He has been named as a highly cited researcher in computer science. He is a Fellow of IEEE.
\end{IEEEbiography}

\vfill

\end{document}